%% file: latex/acl_latex.tex
\documentclass[11pt]{article}
\usepackage[final]{acl}

\usepackage{times}
\usepackage{latexsym}
\usepackage[T1]{fontenc}
\usepackage[utf8]{inputenc}
\usepackage{microtype}
\usepackage{inconsolata}
\usepackage{graphicx}
\usepackage{multirow}
\usepackage{makecell}
\usepackage[table,xcdraw]{xcolor}
\usepackage{amsmath}
\usepackage{amssymb}
\usepackage[normalem]{ulem}
\useunder{\uline}{\ul}{}
\usepackage{tabularx}

\title{Aligning with Your Own Voice: Self-Corrected Preference Learning for Hallucination Mitigation in LVLMs}

\author{
  \textbf{Byeonggeuk Lim}$^{1}$, 
  \textbf{JungMin Yun}$^{2}$, 
  \textbf{JuneHyoung Kwon}$^{2}$, 
  \textbf{Kyeonghyun Kim}$^{2}$,
  \textbf{YoungBin Kim}$^{1,2}$ \\
  $^1$Graduate School of Advanced Imaging Sciences, Multimedia and Film, Chung-Ang University \\
  $^2$Department of Artificial Intelligence, Chung-Ang University \\
  \small{\texttt{\{banggeuk, cocoro357, dirchdmltnv, khyun8072, ybkim85\}@cau.ac.kr}}
}

\begin{document}
\maketitle
\begin{abstract}
Large Vision-Language Models (LVLMs) frequently suffer from hallucinations. Existing preference learning-based approaches largely rely on proprietary models to construct preference datasets. We identify that this reliance introduces a distributional mismatch between the proprietary and target models that hinders efficient alignment. To address this, we propose Alignment via VErified Self-correction DPO (AVES-DPO), a framework that aligns LVLMs using in-distribution data derived from the model's intrinsic knowledge. Our approach employs a consensus-based verification mechanism to diagnose diverse hallucinations and guides the model to self-correct, thereby generating preference pairs strictly compatible with its internal distribution. Extensive experiments demonstrate that AVES-DPO surpasses existing baselines in hallucination mitigation while requiring only 5.2k samples.
\end{abstract}

\section{Introduction}
\label{sec:intro}
\input{latex/01_introduction}

\section{Related Work}
\label{sec:related_work}
\input{latex/02_related_work}

\section{Empirical Analysis of  Distributional Mismatch in Preference Learning}
\label{sec:analysis}
\input{latex/03_analysis}

\section{Method: AVES-DPO}
\label{sec:method}
\input{latex/04_method}

\section{Experiment}
\label{sec:experiment}
\input{latex/05_experiments}

\section{Conclusion}
\label{sec:conclusion}
\input{latex/06_conclusion}

\section*{Limitations}

Despite the effectiveness of AVES-DPO, certain limitations remain that provide avenues for future research. First, our verification of attributes and relationships was strictly constrained to the single-object level to ensure rigorous evaluation, leaving complex scenarios involving multiple co-occurring objects for subsequent investigation. Second, while the 13B model successfully achieved a substantial reduction in hallucination rates, this came at the cost of a slight performance decline on general benchmarks. This phenomenon suggests that as model scale increases, the trade-off between suppressing hallucinations and preserving extensive pre-trained knowledge becomes more pronounced. Consequently, our findings point toward a promising direction for future work in developing scale-aware alignment strategies. Such approaches would aim to mitigate the potential issues inherent in learning from self-generated data while ensuring more comprehensive and robust performance enhancements across models of varying sizes.

\section*{Acknowledgements}
This work was supported by the Institute of Information \& Communications Technology Planning \& Evaluation (IITP) grant funded by the Korea government (MSIT) [RS-2021-II211341, Artificial Intelligence Graduate School Program (Chung-Ang University)] and by the National Research Foundation of Korea (NRF) grant funded by the Korea government (MSIT) (RS-2025-00556246). This research was supported by the "Regional Innovation System \& Education (RISE)" through the Seoul RISE Center, funded by the Ministry of Education (MOE) and the Seoul Metropolitan Government. (2025-RISE-01-024-06).

% \section*{Acknowledgments}

% This document has been adapted
% by Steven Bethard, Ryan Cotterell and Rui Yan
% from the instructions for earlier ACL and NAACL proceedings, including those for
% ACL 2019 by Douwe Kiela and Ivan Vuli\'{c},
% NAACL 2019 by Stephanie Lukin and Alla Roskovskaya,
% ACL 2018 by Shay Cohen, Kevin Gimpel, and Wei Lu,
% NAACL 2018 by Margaret Mitchell and Stephanie Lukin,
% Bib\TeX{} suggestions for (NA)ACL 2017/2018 from Jason Eisner,
% ACL 2017 by Dan Gildea and Min-Yen Kan,
% NAACL 2017 by Margaret Mitchell,
% ACL 2012 by Maggie Li and Michael White,
% ACL 2010 by Jing-Shin Chang and Philipp Koehn,
% ACL 2008 by Johanna D. Moore, Simone Teufel, James Allan, and Sadaoki Furui,
% ACL 2005 by Hwee Tou Ng and Kemal Oflazer,
% ACL 2002 by Eugene Charniak and Dekang Lin,
% and earlier ACL and EACL formats written by several people, including
% John Chen, Henry S. Thompson and Donald Walker.
% Additional elements were taken from the formatting instructions of the \emph{International Joint Conference on Artificial Intelligence} and the \emph{Conference on Computer Vision and Pattern Recognition}.

\bibliographystyle{acl_natbib}
\bibliography{custom}

\input{latex/09_appendix}

\end{document}

%% file: latex/01_introduction.tex
\begin{figure}[t]
    \centering
    \includegraphics[width=1.0\columnwidth]{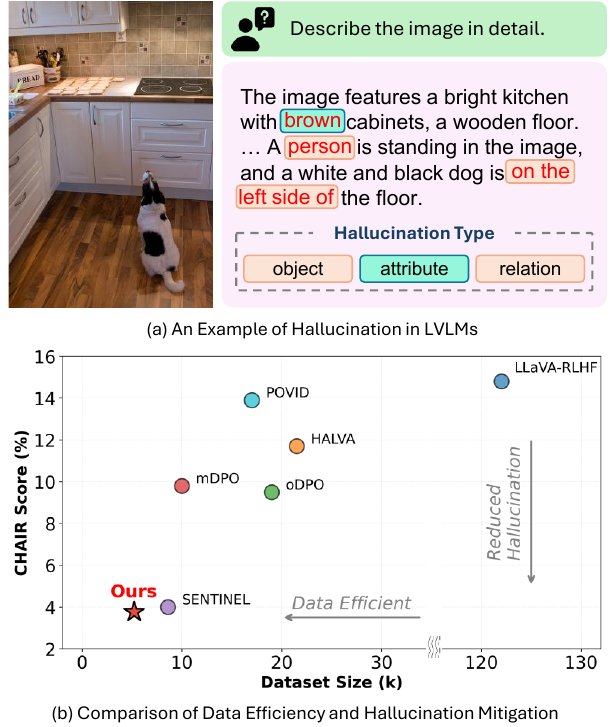}
    \caption{Overview of hallucination types and the effectiveness of the proposed method. (a) An example of hallucinations in LVLMs. (b) Our proposed AVES-DPO achieves the lowest CHAIR score with only 5.2k training samples, demonstrating strong data efficiency.}
    \label{fig1}
\end{figure}

Large Vision-Language Models (LVLMs) have achieved impressive performance across various multimodal tasks~\cite{pmlr-v202-li23q, NEURIPS2023_6dcf277e, Liu_2024_CVPR}. However, existing LVLMs still suffer from hallucination, generating content inconsistent with the input images, such as non-existent objects or misrepresented attributes and relationships as illustrated in Figure~\ref{fig1}(a)~\cite{rohrbach-etal-2018-object, li-etal-2023-evaluating}. To mitigate this issue, preference learning has emerged as a promising approach, which optimizes models to prefer factual responses over hallucinated ones~\cite{xie-etal-2024-v, liu2024mitigating, lee-etal-2024-volcano, sun-etal-2024-aligning}. Among these methods, Direct Preference Optimization (DPO) has become mainstream, enabling efficient policy optimization using offline datasets without explicit reward modeling~\cite{NEURIPS2023_a85b405e}.

Building upon this foundation, recent research has actively applied DPO to enhance the factuality of LVLMs~\cite{Yu_2024_CVPR, Yang_2025_CVPR, sun-etal-2024-aligning}. The primary strategy involves constructing preference datasets where factual responses are favored over hallucinated ones. To achieve this, existing methods typically employ advanced proprietary models, such as GPT-4V~\cite{achiam2023gpt}, to generate high-quality positive samples or to synthesize negative counterparts by manipulating ground-truth captions~\cite{zhao2023beyond, sun-etal-2024-aligning, Yu_2024_CVPR, Yang_2025_CVPR, 10.1007/978-3-031-73116-7_23, Peng_2025_ICCV}. By aligning LVLMs with these constructed preferences, these approaches have successfully demonstrated significant improvements in reducing hallucination rates across various benchmarks.

However, existing DPO-based approaches face two key limitations. First, constructing preference pairs relies heavily on external supervision from proprietary models such as GPT-4V~\cite{sun-etal-2024-aligning, Yang_2025_CVPR, sarkar2025mitigating}. This dependence introduces a distribution mismatch, as the generation patterns of these external models often diverge from the target model's intrinsic distribution, thereby hindering effective preference alignment. Second, existing datasets and training objectives are disproportionately biased towards object hallucinations, which primarily concern determining object presence~\cite{Peng_2025_ICCV, Park_2025_CVPR, he-etal-2025-evaluating}. Although subtle hallucinations, such as misrepresented attributes or relationships, current methods remain limited in addressing these diverse and fine-grained error types. To address these challenges, we explore the self-correction strategy on diverse hallucination types, mitigating the distribution mismatch.

In this paper, we introduce \textbf{A}lignment via \textbf{VE}rified \textbf{S}elf-correction DPO (AVES-DPO), a framework comprising two distinct stages: Hallucination Verification and Self-correction. In the verification stage, generated responses are scrutinized across object, attribute, and relationship levels to identify whether specific elements are hallucinated. Subsequently, the LVLMs self-refine their outputs based on these diagnoses, rectifying errors while enriching the response with missing visual details. Using these self-refined outputs as preferred data provides in-distribution training signals, effectively mitigating the distribution mismatch. Furthermore, this comprehensive process also overcomes the limitations of targeting only single-type hallucinations. Finally, we employ these refined outputs to construct preference pairs for alignment training, demonstrating that this approach yields more efficient and effective learning compared to models trained with responses generated by proprietary models.

Extensive evaluations across various hallucination benchmarks demonstrate that our method significantly outperforms existing baselines. Notably, the 7B model achieves state-of-the-art (SOTA) performance on over 60\% of these diverse benchmarks. Through this framework, as illustrated in Figure~\ref{fig1}(b), AVES-DPO achieves superior performance with only 5.2k samples, representing approximately 25 times less data than LLaVA-RLHF, highlighting its exceptional data efficiency.

%% file: latex/02_related_work.tex
%Hallucination in LVLMs refers to the phenomenon where generated text is inconsistent with the provided visual evidence or user instructions.
%Hallucination in LVLMs refers to the phenomenon where text generated by the model is inconsistent with visual evidence or user instructions. Existing research to mitigate this can be broadly categorized into three streams depending on the approach.

% [문단별 피드백]
% 1) 첫 문장 조금 더 구체적으로. hallucination 완화를 위해서 어떤 데이터 확보라던가 
% 2) 접근법의 문제 지적 전에 기존 연구 조금만 더 추가하는 게 좋을 것 같습니다. 접근법에 대한 문제 지적이 너무 바로 들어가는 느낌이라 기존 연구 조금만 더 보충 
% One line of research on hallucination mitigation focuses on directly optimizing model parameters using high-quality training data. While early studies constructed human-annotated datasets for SFT, the substantial costs of securing large-scale annotations motivated a shift toward preference learning, which optimizes models by contrasting preferred and dispreferred responses.
\noindent \textbf{Learning from Feedback.} Early efforts to mitigate hallucination primarily relied on Supervised Fine-Tuning (SFT) with correction signals derived from proprietary LLMs. For instance, LRV-Instruction and ReCaption leverage proprietary LLMs like GPT-4 to generate counterfactual visual instructions or rewrite captions, training models to suppress hallucinations ranging from coarse object presence to fine-grained details ~\cite{liu2024mitigating, 10.1007/978-3-031-53302-0_3}.

Recent research has expanded toward preference learning to optimize models based on the divergence between positive and negative responses ~\cite{NEURIPS2023_a85b405e}. LLaVA-RLHF employs Proximal Policy Optimization (PPO)-based reinforcement learning, mitigating hallucinations by incorporating factual information into the reward model ~\cite{schulman2017proximal, sun-etal-2024-aligning}. However, to address the inherent training instability and complexity of PPO, various DPO-based methods have subsequently been proposed. For instance, POVID leverages GPT-4V to inject plausible hallucinations into ground-truth data and generate dispreferred responses via distorted images ~\cite{zhou2024aligning}. To handle modality-specific challenges, mDPO ~\cite{wang-etal-2024-mdpo} introduces an image preference objective to address the unconditional preference problem where the model ignores visual inputs and relies solely on text, whereas oDPO utilizes object masks for fine-grained, object-level optimization ~\cite{he-etal-2025-evaluating}. Distinct from the DPO paradigm, HALVA adopts a contrastive learning approach with generative data augmentation, offering an efficient alternative for mitigating hallucinations while preserving general model capabilities ~\cite{sarkar2025mitigating}.

Despite these advances, most methods depend on proprietary models such as GPT-4V or Gemini Vision Pro ~\cite{team2023gemini} to construct preference pairs, incurring substantial costs and potentially causing distribution mismatch between the external teacher and target models, which may degrade training efficiency and generalization performance ~\cite{zhou2024aligning, compagnoni2025mitigating, Yang_2025_CVPR, sarkar2025mitigating}.

\noindent \textbf{Inference-time Mitigation.} Another line of research controls hallucinations at inference time without modifying model parameters, primarily through post-hoc verification. While some methods like LURE focus on rectifying hallucinations based on statistical cues and uncertainty ~\cite{zhou2024analyzing}, recent work has increasingly adopted a decompose-and-verify paradigm using external experts. In this vein, Woodpecker validates generated key concepts using visual experts (e.g., object detectors) ~\cite{yin2024woodpecker}, a process further structured by Pelican, which decomposes claims into sub-claims to execute Program-of-Thought verification via composable tools ~\cite{sahu-etal-2024-pelican}. % 가능하면 마지막 문장 빼고 관련 연구 보완

Another line of research focuses on inference time intervention through modified decoding strategies. VCD applies contrastive decoding by leveraging logit differences between original and distorted images ~\cite{Leng_2024_CVPR}. OPERA penalizes over-trust attention patterns and reallocates decoding when knowledge aggregation causes the model to neglect image tokens ~\cite{Huang_2024_CVPR}. However, these methods exhibit sensitivity to hyperparameter settings, limiting their generalizability across diverse domains. Furthermore, they predominantly focus on object existence verification, leaving fine-grained hallucinations involving attributes and relations inadequately addressed.

%% file: latex/03_analysis.tex
\begin{figure}[t]
    \centering
    \includegraphics[width=1.0\columnwidth]{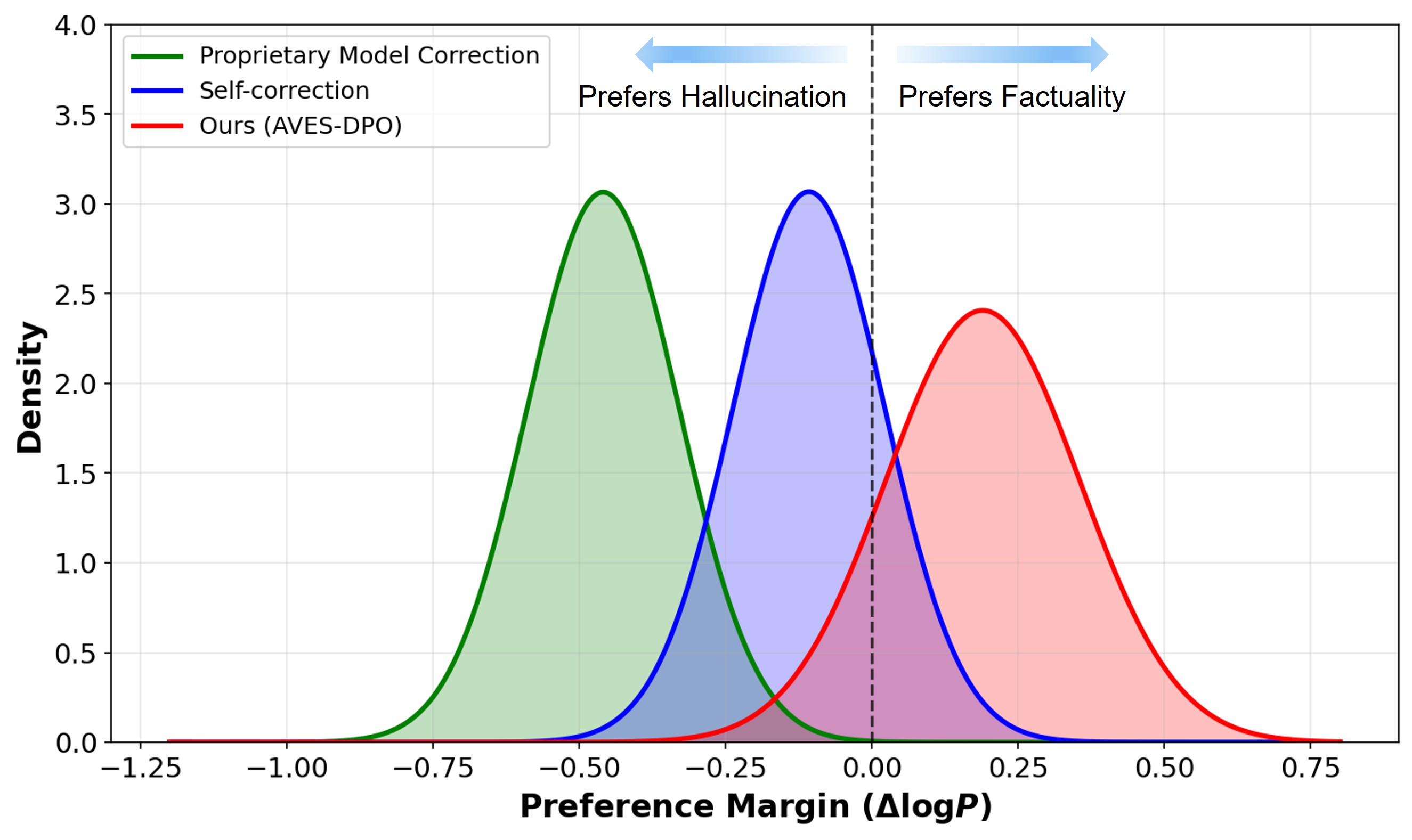}
    \caption{Distributional alignment of proprietary model vs self-correction. Unlike proprietary model correction (green) which suffers from distribution mismatch, self-correction (blue) aligns with the model’s internal distribution. Our approach (red) successfully shifts the preference margin toward factuality.}
    \label{fig2}
\end{figure}
 
% [문단별 피드백]
% 1) 어떤 데이터셋, 어떤 모델에 대해서 수행했는지 언급 추가 
% 2) 각 notation이 뭔지 설명 추가, preference margin이 어떤 의미고, 뭘 측정하는지 명시 
% 3) Proprietary Model Correction, Self-correction이 각각 뭔지 구체적으로 명시
We conduct a preliminary empirical analysis to investigate the distribution mismatch issue arising from reliance on external models. This pilot study provides the rationale for adopting a self-correction mechanism.

\paragraph{Settings.}
We conduct experiments using LLaVA-1.5-7B on 200 hallucinated samples identified from the COCO dataset~\cite{10.1007/978-3-319-10602-1_48}. We generate responses using the prompt ``Describe the image in detail.'' and employ our hallucination verification module to filter samples, resulting in initial responses $y_{init}$ containing verified hallucinations.

To investigate the distributional discrepancy arising from different correction sources, we prepare two types of corrected responses $y_{corr}$. For Proprietary Model Correction, we instruct GPT-4V to correct and remove hallucinations and subsequently enrich the description, generating corrections external to the target model's distribution. For Self-correction, LVLMs themselves rectify hallucinations and enrich the responses using hallucination information derived from our verification framework. We then measure the preference margin $\mathcal{M}$ to quantify the distributional shift, defined as the difference in response-averaged log probabilities $\frac{1}{L} \sum_{i} \log P_\theta(y_i|x, y_{<i})$ between the corrected response $y_{corr}$ and the initial response $y_{init}$.

\paragraph{Results.}
Figure~\ref{fig2} illustrates the distribution of these preference margins. Proprietary Model Correction results in a relatively large negative margin, indicating that significant divergence between the proprietary model's generation patterns and the target model's intrinsic distribution. Learning from such data forces the model to adapt to distinct stylistic patterns, potentially leading to inefficient optimization and limited generalization. In contrast, Self-correction exhibits a margin distribution centered near zero. This observation confirms that corrections generated by the model itself reside within its internal distribution, providing natural and compatible learning signals. These findings suggest that utilizing self-corrected data minimizes distributional discrepancy, offering a more effective foundation for preference alignment compared to external supervision. This insight motivates the design of our proposed framework, AVES-DPO, which is detailed in the following section.

%Building on these observations, our proposed AVES-DPO demonstrates a distinct shift of the preference margin toward positive value. This confirms that our method successfully aligns the model to assign higher probabilities to factual responses over hallucinated ones. These findings suggest that utilizing self-corrected data provides compatible learning signals that minimize distributional discrepancy, offering a more effective strategy for preference alignment.

%% file: latex/04_method.tex
\begin{figure*}[t]
    \centering
    \includegraphics[width=\textwidth]{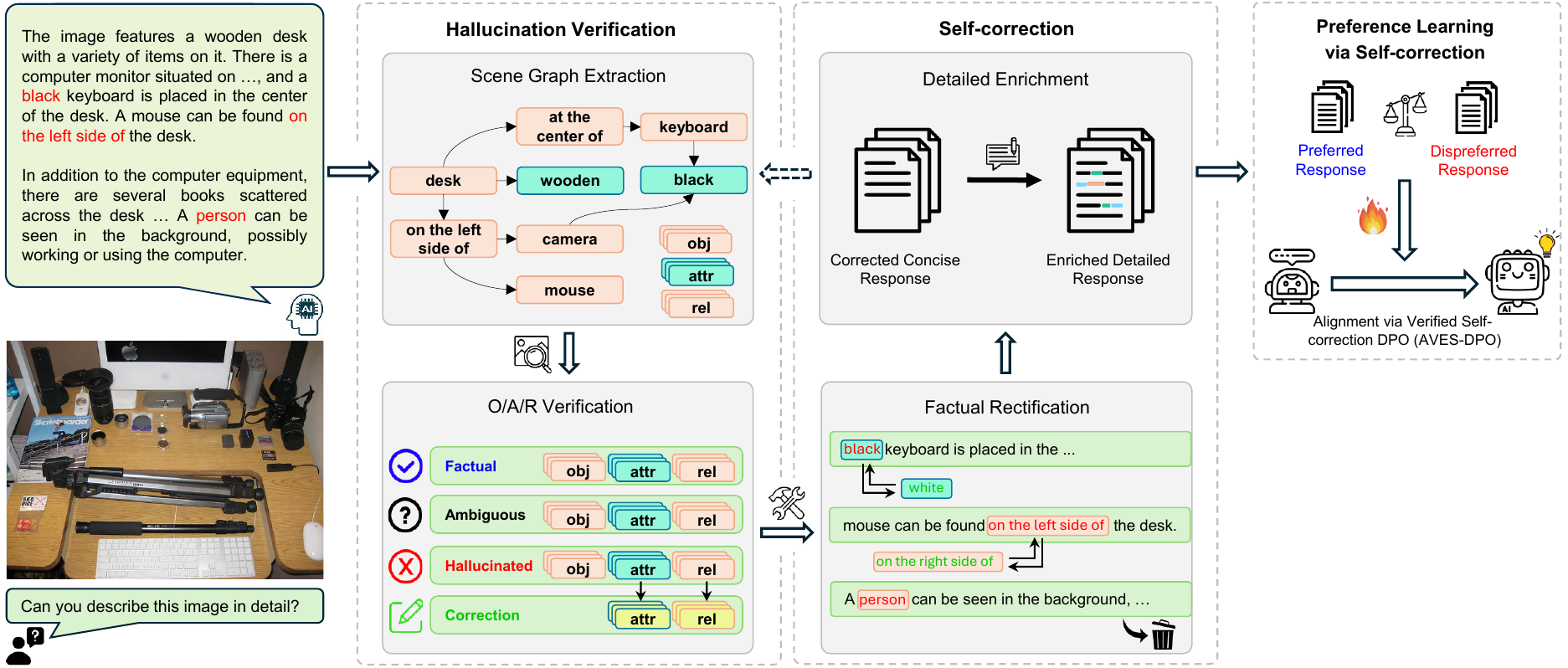}
    \caption{The overall framework of the proposed AVES-DPO.}
    \label{fig3}
\end{figure*}

\subsection{Preliminaries}

\paragraph{Large Vision-Language Models.} Typically, LVLMs comprise three core components: a visual encoder that processes visual inputs, a modality alignment module that matches dimensions between modalities, and a LLM responsible for text generation. Given an input image $X_v$ and text instruction $X_t$, the visual encoder converts the image into visual features $H_v$. These features are projected via the alignment module into visual tokens $H_v'$, which are mapped to the embedding space of the LLM. The LLM then takes the projected visual tokens $H_v'$ and the tokenized text instructions $X_t'$ as a combined input $x=\{H_v', X_t'\}$ to generate a response $y$ in an auto-regressive manner, modeling the conditional probability distribution $\pi_\theta(y|x)$.

\vspace{2mm}
\paragraph{Direct Preference Optimization.} To align LVLMs with human preferences, we adopt DPO, which directly optimizes the policy using preference data without requiring an explicit reward model. While traditional RLHF approaches require training a separate reward model, DPO derives the optimal policy by expressing the implicit reward function $r(x,y)$ in terms of the policy $\pi_\theta$ and a reference model $\pi_{ref}$:

\begin{equation}
    r(x,y)=\beta \log \frac{\pi_\theta(y|x)}{\pi_{ref}(y|x)} + \beta \log Z(x).
    \label{eq:reward_model}
\end{equation}

\noindent Here, $x$ denotes the multimodal input, $Z(x)$ represents the partition function, and $\beta$ is a coefficient controlling the deviation. Under the Bradley-Terry model~\cite{19ff28b9-64f9-3656-ba40-08326a05748e}, we can obtain the DPO loss function that directly optimizes the policy using a preference dataset $\mathcal{D}=\{x, y_w, y_l\}$:

\begin{equation}
\begin{split}
    & \mathcal{L}_{DPO}(\pi_\theta;\pi_{ref}) \\
    & = -\mathbb{E}_{(x, y_w, y_l) \sim \mathcal{D}} \left[ \log \sigma \left( r(x, y_w) - r(x, y_l) \right) \right] \\
    & = -\mathbb{E}_{(x, y_w, y_l) \sim \mathcal{D}} \left[ \log \sigma \left( \beta \log \frac{\pi_\theta(y_w|x)}{\pi_{ref}(y_w|x)} \right. \right. \\
    & \quad \quad \quad \quad \quad \quad \quad \quad \quad \left. \left. - \beta \log \frac{\pi_\theta(y_l|x)}{\pi_{ref}(y_l|x)}. \right) \right]
\end{split}
\label{eq:dpo_loss}
\end{equation}

\noindent In this study, we utilize Eq.~(\ref{eq:dpo_loss}) to mitigate hallucinations by optimizing the likelihood difference between the model's self-corrected response $y_w$ and the initial hallucinated response $y_l$.

\subsection{Construction of Self-corrected Data}
\label{sec:4.2}

To construct a high-quality preference dataset, we propose a two-stage framework: Hallucination Verification and Self-correction. As illustrated in Figure~\ref{fig3}, the Hallucination Verification stage consists of Scene Graph Extraction and O/A/R Verification, while the Self-correction stage consists of Factual Rectification and Detailed Enrichment.

\subsubsection{Hallucination Verification}
\label{subsec:4.2.1}

\paragraph{Scene Graph Extraction.} To analyze the initial response $y$, we convert it into a scene graph $G=(O,A,R)$, comprising a set of objects $O=\{o_i\}$, attributes $A=\{a_{ij}\}$, and relationships $R=\{r_{ijk}\}$. This structured representation enables fine-grained hallucination detection. For precise correction, we utilize a vocabulary derived from the GQA dataset, organized into semantic subtypes to efficiently retrieve appropriate replacement candidates within the same category (details provided in Appendix ~\ref{appendixA.1}). The vocabulary is organized into semantic subtypes to enable the efficient retrieval of appropriate correction candidates within the same category when a hallucination is detected.

\paragraph{O/A/R Verification.}
To verify the factuality of each extracted element, we adopt a consensus-based verification strategy utilizing two independent verification models, $V_1$ and $V_2$. To ensure reliability, a label is finalized only when both models agree; otherwise, it is classified as \textit{Ambiguous}. The verification result $L(x)$ for an arbitrary element $x \in O \cup A \cup R$ is defined as follows:
\begin{equation}
L(x) =
\begin{cases}
l & \text{if } V_1(x) = V_2(x) = l, \\
\text{\textit{Ambiguous}} & \text{otherwise},
\end{cases}
\end{equation}
where $l \in \{\text{\textit{Factual}}, \text{\textit{Hallucinated}}\}$.

\textit{(i) Object Verification.} First, we assess the presence of objects within the image. As the primary verifiers $(V_1, V_2)$, we employ YOLO ~\cite{Cheng_2024_CVPR} and Grounding DINO ~\cite{10.1007/978-3-031-72970-6_3}, known for their robust object detection. Given the potential limitations of detection models, objects initially classified as \textit{Ambiguous} undergo a secondary verification phase using the Qwen3-VL series\footnote{Qwen3-VL-30B-A3B-Instruct, Qwen3-VL-32B-Instruct} ~\cite{bai2025qwen3vltechnicalreport}. In this stage, the label $L(o_i)$ is updated to the consensus label (i.e., \textit{Factual} or \textit{Hallucinated}) only if both LVLM models independently produce the identical output.

\textit{(ii) Attribute \& Relation Verification.} Attributes and relationships are verified exclusively for objects confirmed as $L(o_i)=\text{\textit{Factual}}$. We utilize the two models from the Qwen3-VL series as verifiers under the same consensus rule. If an element is classified as \textit{Hallucinated}, the system searches the pre-constructed vocabulary for a correct attribute or relationship matching the element's semantic subtype to serve as a correction candidate.

\subsubsection{Self-correction}
\label{subsec:4.2.2}

Guided by the diagnoses from the Hallucination Verification stage, Self-correction generates the final preferred response through two phases: Factual Rectification and Detailed Enrichment.

\noindent \textbf{Factual Rectification.}
This phase ensures factuality by reconstructing the text based on verification results. Specifically, elements identified as hallucinations are either removed or replaced with correct content. This process yields a concise, factually accurate response free from hallucinated content.

\input{latex/table/table1}
\input{latex/table/table2}
\noindent \textbf{Detailed Enrichment.}
To prevent oversimplification caused by mere error removal, this phase restores descriptive capability by incorporating missing visual details. Given the risk of introducing new hallucinations during generation, we employ a cyclic iterative filtering pipeline. Specifically, the enriched text is fed back into the Hallucination Verification module. If no hallucinations are detected, the response passes; if hallucinations are found, it is sent back to the Self-Correction stage for re-generation. We repeat this verification-correction loop up to 3 times for the 7B model and 5 times for the 13B model. Samples failing to yield a valid response within these attempts are discarded to guarantee high data quality. This rigorous process produces a final preferred response $y^+$ that balances factual accuracy with visual richness, which is then paired with the initial hallucinated response $y^-$ for Preference Learning.

\subsection{Preference Learning via Self-correction}

Following the procedures in Section~\ref{sec:4.2}, we construct a dataset $\mathcal{D}_{SC}$ comprising pairs of the model's initial hallucinated responses and enriched response obtained through precise refinement. Each sample is defined as $(x, y^+, y^-)$, consisting of the input image and text $x=\{X_v, X_t\}$, the preferred response $y^+$, and the dispreferred response $y^-$. Here, $y^-$ corresponds to the initial response identified as hallucinated during the verification process in Section~\ref{subsec:4.2.1}, while $y^+$ denotes the final enriched response generated via the Self-correction process in Section~\ref{subsec:4.2.2}.

Our training objective is to induce the model, conditioned on input $x$, to maximize the likelihood of generating the rectified response $y^+$, while minimizing the probability of generating the hallucinated response $y^-$. To achieve this, we employ the DPO loss function. Based on this objective, our proposed AVES-DPO loss is formulated as follows:%We term our approach \textbf{A}lignment via \textbf{VE}rified \textbf{S}elf-correction DPO (AVES-DPO), and formulate it as follows:
\begin{equation}
\begin{split}
    & \mathcal{L}_{\text{AVES-DPO}}(\pi_\theta;\pi_{ref}) \\
    & = -\mathbb{E}_{(x, y^+, y^-) \sim \mathcal{D}_{SC}} \left[ \log \sigma \left( \beta \log \frac{\pi_\theta(y^+|x)}{\pi_{ref}(y^+|x)} \right. \right. \\
    & \quad \quad \quad \quad \quad \quad \quad \quad \quad \left. \left. - \beta \log \frac{\pi_\theta(y^-|x)}{\pi_{ref}(y^-|x)} \right) \right].
\end{split}
\label{eq:aves_dpo_loss}
\end{equation}

%% file: latex/table/table1.tex
\begin{table*}[t!]
\centering
\renewcommand{\arraystretch}{1.3}
\resizebox{1.0\textwidth}{!}{%
\begin{tabular}{lcc|cc|cccccc|cc}
\Xhline{3\arrayrulewidth}
\rowcolor[HTML]{EFEFEF} 
\multicolumn{1}{c}{\cellcolor[HTML]{EFEFEF}} 
& \cellcolor[HTML]{EFEFEF} 
& \cellcolor[HTML]{EFEFEF} 
& \multicolumn{2}{c|}{\cellcolor[HTML]{EFEFEF}\textbf{Object-Hal}} 
& \multicolumn{6}{c|}{\cellcolor[HTML]{EFEFEF}\textbf{AMBER}} 
& \multicolumn{2}{c}{\cellcolor[HTML]{EFEFEF}\textbf{MMHal-Bench}} \\ \cline{4-13}

\rowcolor[HTML]{EFEFEF} 
\multicolumn{1}{c}{\multirow{-2}{*}{\cellcolor[HTML]{EFEFEF}\textbf{Method}}} 
& \multirow{-2}{*}{\cellcolor[HTML]{EFEFEF}\textbf{Data Size}} 
& \multirow{-2}{*}{\cellcolor[HTML]{EFEFEF}\textbf{Feedback}} 
& \textbf{CHAIR$_S$$\downarrow$} 
& \textbf{CHAIR$_I$$\downarrow$} 
& \textbf{CHAIR$\downarrow$} 
& \textbf{Cover$\uparrow$} 
& \textbf{Hal Rate$\downarrow$} 
& \textbf{Cog$\downarrow$} 
& \textbf{Acc$\uparrow$} 
& \textbf{F1$\uparrow$} 
& \textbf{Score$\uparrow$} 
& \textbf{Hal Rate$\downarrow$} \\ \hline\hline

\textit{LLaVA-1.5-7B$^{\dagger}$} & - & - & 51.4 & 14.8 & 7.1 & 50.5 & 32.5 & 3.8 & 78.8 & 83.0 & 2.22 & 0.57 \\
+ VCD$^{\S}$ ~\cite{Leng_2024_CVPR} & - & - & 53.3 & 15.7 & 6.9 & 50.6 & 32.2 & 3.7 & 72.0 & 74.8 & 2.12 & 0.54 \\
+ LLaVA-RLHF$^{\dagger}$ ~\cite{sun-etal-2024-aligning} & 122k & Self-Reward & 53.6 & 14.8 & 8.4 & 52.3 & 42.5 & 4.6 & 70.2 & 76.5 & 2.06 & 0.64 \\
+ HALVA$^{\ddagger}$ ~\cite{sarkar2025mitigating} & 21.5k & GPT-4V & 41.4 & 11.7 & 6.6 & 53.0 & 32.2 & 3.4 & - & 83.4 & 2.25 & 0.54 \\
+ POVID$^{\dagger}$ ~\cite{zhou2024aligning} & 17k & GPT-4V & 45.8 & 13.9 & 7.1 & 50.2 & 31.5 & 3.6 & 78.4 & 82.8 & 2.18 & 0.57 \\
+ oDPO$^{\S}$ ~\cite{he-etal-2025-evaluating} & 19k & GPT-4V & 34.3 & 9.5 & 4.6 & \textbf{53.4} & 25.1 & 2.4 & 80.2 & 84.1 & \textbf{2.50} & \textbf{0.49} \\
+ mDPO$^{*}$ ~\cite{wang-etal-2024-mdpo} & 10k & GPT-4V & 35.7 & 9.8 & 4.4 & 52.4 & 24.5 & 2.4 & - & - & 2.39 & 0.54 \\
+ SENTINEL$^{\dagger}$ ~\cite{Peng_2025_ICCV} & 8.6k & Self & 12.6 & 4.0 & \textbf{2.9} & 43.7 & 14.6 & 1.2 & 80.3 & 85.2 & 2.04 & 0.58 \\
\rowcolor[HTML]{EEF3FF} 
\textbf{+ AVES-DPO (Ours)} & 5.2k & Self & \textbf{12.2} & \textbf{3.9} & 3.3 & 40.8 & \textbf{12.6} & \textbf{0.9} & \textbf{82.3} & \textbf{86.9} & 2.35 & 0.53 \\ \hline

\textit{LLaVA-1.5-13B$^{\dagger}$} & - & - & 50.4 & 14.3 & 6.7 & 51.3 & 31.0 & 3.5 & 80.4 & 83.7 & 2.30 & 0.54 \\
+ VCD$^{\S}$ ~\cite{Leng_2024_CVPR} & - & - & 47.7 & 13.2 & 6.7 & 51.3 & 31.0 & 3.5 & 71.5 & 73.5 & 2.40 & 0.51 \\
+ LLaVA-RLHF$^{\dagger}$ ~\cite{sun-etal-2024-aligning} & 122k & Self-Reward & 47.0 & 12.4 & 6.9 & 51.7 & 35.7 & 3.3 & 81.0 & 86.4 & 2.14 & 0.66 \\
+ HALVA$^{\ddagger}$ ~\cite{sarkar2025mitigating} & 21.5k & GPT-4V & 45.4 & 12.8 & 6.4 & \textbf{52.6} & 30.4 & 3.2 & - & \textbf{86.5} & \textbf{2.58} & \textbf{0.45} \\
+ oDPO$^{\S}$ ~\cite{he-etal-2025-evaluating} & 19k & GPT-4V & 34.7 & 9.8 & 4.3 & 52.1 & 23.1 & 2.2 & 79.3 & 82.2 & 2.74 & \textbf{0.45} \\
+ SENTINEL$^{\dagger}$ ~\cite{Peng_2025_ICCV} & 7k & Self & 11.8 & 3.3 & \textbf{2.8} & 45.3 & \textbf{13.2} & 1.0 & \textbf{81.4} & 84.7 & 2.48 & 0.49 \\
\rowcolor[HTML]{EEF3FF} 
\textbf{+ AVES-DPO (Ours)} & 4.6k & Self & \textbf{11.3} & \textbf{3.0} & 3.9 & 40.4 & 17.6 & \textbf{0.9} & 80.4 & 83.3 & 2.36 & 0.58 \\
\Xhline{3\arrayrulewidth}
\end{tabular}
}
\caption{Comparison of hallucination mitigation performance across various benchmarks. For baseline algorithms with available official checkpoints, we re-evaluate the models, and these results are marked with $^{\dagger}$.
Results sourced from ~\cite{he-etal-2025-evaluating} are denoted by $^{\S}$, \cite{sarkar2025mitigating} by $^{\ddagger}$, and ~\cite{wang-etal-2024-mdpo} by $^{*}$.}
\label{tab:tab1_main}
\end{table*}

%% file: latex/table/table2.tex
\begin{table*}[t!]
\centering
\renewcommand{\arraystretch}{1.3}
\resizebox{0.8\textwidth}{!}{%
\begin{tabular}{l|cccccc|c}
\Xhline{3\arrayrulewidth}

% === Header Row ===
\rowcolor[HTML]{EFEFEF}
\multicolumn{1}{c|}{\textbf{Method}} &
\textbf{Existence} & \textbf{Attribute} & \textbf{State} &
\textbf{Number} & \textbf{Action} & \textbf{Relation} & \textbf{Overall} \\
\hline\hline

% === 7B Models ===
\textit{LLaVA-1.5-7B} & 87.4 & 77.7 & 75.3 & 80.7 & 84.2 & 58.5 & 78.8 \\
+ LLaVA-RLHF ~\cite{sun-etal-2024-aligning} & 67.4 & 73.2 & 72.0 & 72.0 & 83.8 & {\ul 64.7} & 70.2 \\
+ POVID ~\cite{zhou2024aligning} & 87.8 & 77.3 & 75.4 & 79.1 & 84.1 & 55.5 & 78.4 \\
+ SENTINEL ~\cite{Peng_2025_ICCV} & \textbf{91.4} & {\ul 80.0} & {\ul 77.4} & \textbf{82.9} & \textbf{87.6} & 48.7 & {\ul 80.3} \\
\rowcolor[HTML]{EEF3FF} 
\textbf{+ AVES-DPO (Ours)} & {\ul 90.6} & \textbf{80.4} & \textbf{79.1} & {\ul 81.0} & {\ul 87.0} & \textbf{66.6} & \textbf{82.3} \\ \hline

% === 13B Models ===
\textit{LLaVA-1.5-13B} & 83.8 & {\ul 81.9} & \textbf{80.0} & {\ul 84.8} & {\ul 85.7} & 63.0 & 80.4 \\
+ LLaVA-RLHF ~\cite{sun-etal-2024-aligning} & \textbf{93.8} & 77.1 & 73.7 & 81.8 & 85.4 & 60.5 & {\ul 81.0} \\
+ SENTINEL ~\cite{Peng_2025_ICCV} & {\ul 85.5} & \textbf{82.2} & {\ul 79.9} & \textbf{86.1} & 85.5 & {\ul 65.6} & \textbf{81.4} \\
\rowcolor[HTML]{EEF3FF} 
\textbf{+ AVES-DPO (Ours)} & 82.6 & 79.6 & 76.0 & {\ul 84.8} & \textbf{87.8} & \textbf{77.5} & 80.4 \\
\Xhline{3\arrayrulewidth}

\end{tabular}
}
\caption{Detailed evaluation results on the discriminative tasks of the AMBER benchmark.}
\label{tab2}
\end{table*}

%% file: latex/05_experiments.tex
\input{latex/table/table3}

\subsection{Experimental Setup}

\noindent \textbf{Models and datasets.}
To evaluate the effectiveness of the proposed method, we employ two representative LVLMs with differing parameter scales: LLaVA-1.5-7B and LLaVA-1.5-13B ~\cite{Liu_2024_CVPR}. Both models utilize CLIP-ViT-L-336px ~\cite{pmlr-v139-radford21a} as the visual encoder, with Vicuna-7B and Vicuna-13B ~\cite{chiang2023vicuna} as the respective language models. Using MS-COCO train set as the image source, we construct 5.2k and 4.6k preference pairs for the 7B and 13B models, respectively.

\noindent \textbf{Evaluation metrics.}
We evaluate on four benchmarks: Object HalBench ~\cite{rohrbach-etal-2018-object} for standard object hallucination metrics, AMBER ~\cite{wang2023amber} for comprehensive analysis covering both generative and discriminative tasks, MMHal-Bench ~\cite{sun-etal-2024-aligning} for GPT-4-based quality assessment in open-ended evaluations, and the MME Benchmark ~\cite{fu2025mme} to assess multimodal generalization capabilities. Details are provided in the Appendix ~\ref{appendixC}.

\noindent \textbf{Baselines.}
We compare our proposed method against several state-of-the-art (SOTA) techniques. Specifically, we selected VCD~\cite{Leng_2024_CVPR} as a representative method for decoding strategy intervention. Additionally, we included LLaVA-RLHF~\cite{sun-etal-2024-aligning}, HALVA~\cite{sarkar2025mitigating}, oDPO~\cite{he-etal-2025-evaluating}, mDPO~\cite{wang-etal-2024-mdpo}, and SENTINEL~\cite{Peng_2025_ICCV} as comparative baselines, representing prominent approaches in preference optimization.

\noindent \textbf{Implementation details.}
We apply Low-Rank Adaptation (LoRA) and use AdamW as the optimizer. For both the 7B and 13B models, we set the LoRA rank to 128 and alpha to 256, training for 1 epoch. The parameter $\beta$ in AVES-DPO, which controls the deviation from the reference model, is set to 0.1. All experiments are conducted on a single NVIDIA RTX A6000 Ada GPU. Further details are provided in the Appendix~\ref{appendixB}.

\input{latex/table/table4}

\subsection{Main Results}

\noindent \textbf{Findings 1: Superior Performance with High Data Efficiency.}
Table ~\ref{tab:tab1_main} presents results on three standard hallucination benchmarks. Despite utilizing a smaller dataset than baseline methods, AVES-DPO demonstrates strong performance across multiple evaluation dimensions. For the 7B model, AVES-DPO achieves the best performance on 60\% of the evaluation metrics. For the 13B model, it achieves the lowest Object-Hal scores with only 4.6k training samples. These results validate that AVES-DPO eliminates the dependency on proprietary API calls while achieving superior performance with minimal data, making it a practical and scalable solution for hallucination mitigation.

\noindent \textbf{Findings 2: Improvements in Fine-grained Performance, especially in relation.}
Table~\ref{tab2} presents the fine-grained evaluation results on the AMBER discriminative task. For the 7B model, AVES-DPO demonstrates consistently superior performances, particularly achieving substantial gains in the relation category where existing methods often struggle. While SENTINEL shows performance degradation compared to the base model, AVES-DPO achieves a robust score of 66.6. For the 13B model, although overall score remains competitive, our method achieves the highest scores in relation category with a score of 77.5, mirroring the trend observed in the 7B model. These consistent improvements in relational reasoning across both model scales verify that our proposed verification and self-correction framework effectively captures complex inter-object relationships.

\noindent \textbf{Findings 3: Enhancement of General Multimodal Capabilities.} To assess generalization capability across diverse domains, we evaluate on the MME Benchmark, which comprises various subtypes. As shown in Table ~\ref{tab3}, for the 7B model, AVES-DPO achieves an overall score of 1491.1, demonstrating a significant performance improvement compared to existing methods. This result confirms that our framework not only mitigates hallucinations but also enhances fundamental multimodal perception abilities. However, we observed a performance trade-off in the 13B model. Given that MME includes subtasks heavily reliant on external knowledge, this pattern suggests that the 13B model becomes more conservative during the hallucination suppression, prioritizing factual grounding over broad knowledge retrieval. This highlights a challenge in balancing the hallucination mitigation with the preservation of extensive multimodal capabilities when scaling to larger models.

\begin{figure}[t!]
    \centering
    \includegraphics[width=0.9\linewidth]{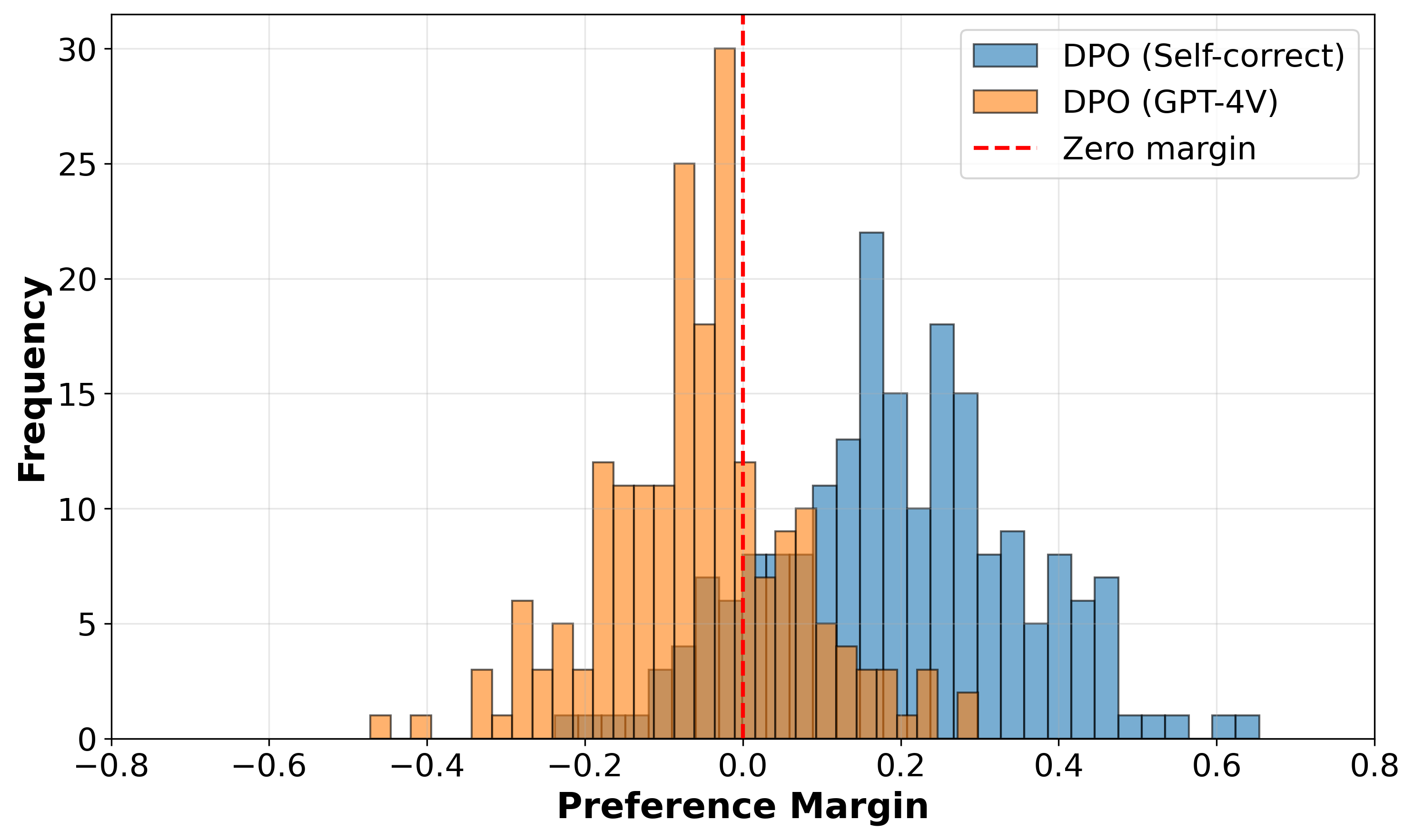}
    \caption{Distribution of preference margins. External GPT-4V supervision (orange) results in margins centered near zero due to distribution mismatch. In contrast, our approach (blue) leverages in-distribution data to induce a significant positive shift.}
    \label{fig:last}
\end{figure}
\input{latex/table/table16}

\begin{figure}[t!]
    \centering
    \includegraphics[width=1.0\columnwidth]{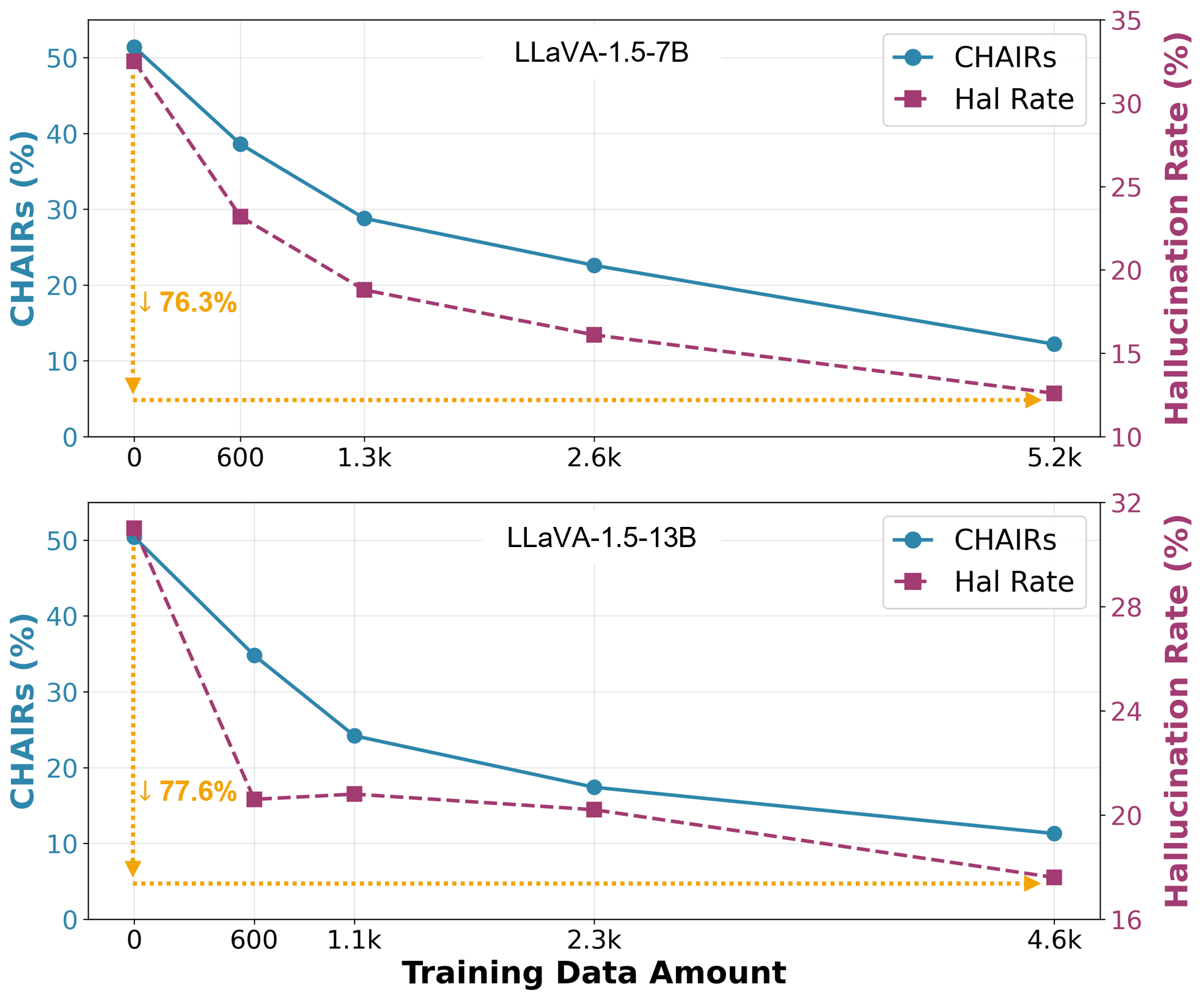}
    \caption{Impact of training data size on hallucination mitigation performance.}
    \label{fig4}
\end{figure}

\subsection{Ablation Study and Analysis}

\noindent \textbf{Robustness of verification strategy.} To assess the efficacy and robustness of our Hallucination Verification process, we conduct experiments on LLaVA-1.5-7B, LLaVA-1.5-13B, and GPT-4V. Each model detects and rectifies hallucinations in initial responses, and we measure the resulting reduction in hallucination rates. As shown in Table~\ref{tab4}, our approach consistently reduces both sentence- and object-level hallucination rates across all models. Notably, LLaVA-1.5-13B achieves relative reductions of 49.6\% and 46.9\% in CHAIRs and CHAIRi, respectively. Furthermore, even for GPT-4V, which already exhibits relatively low hallucination rates, our verification strategy yields additional gains. These results demonstrate that our verification strategy is model-agnostic and robust across varying model capacities.

\noindent \textbf{Effect of distribution mismatch mitigation.} To demonstrate the impact of utilizing in-distribution data, we train a comparative model following the setup in Section~\ref{sec:analysis}. Specifically, we construct an additional baseline dataset of 5.2k samples using responses generated by proprietary models. We then visualize the preference margin $\mathcal{M}$ to quantify the distributional alignment of each correction source. As shown in Figure~\ref{fig:last}, the distribution for proprietary models concentrates predominantly in the negative region. In contrast, our self-correction approach exhibits a significant positive shift, demonstrating that LVLMs can successfully align with preference signals within their own distribution. This disparity translates into empirical performance: as shown in Table~\ref{tab16}, supervision from proprietary models yields only marginal improvement, while our in-distribution training achieves a significant reduction in hallucination rates. These results confirm that leveraging in-distribution data enables more effective preference alignment than external supervision. This finding is further supported by the preference margin distribution of AVES-DPO in Figure~\ref{fig2}, which shifts into the positive region, indicating that the model successfully learns to prefer factual responses.

%These results confirm that leveraging in-distribution data enables more effective preference alignment for LVLMs than external supervision.}

\noindent \textbf{Effect of data scale.} To validate the data efficiency of AVES-DPO, we analyze performance across varying training data sizes, as illustrated in Figure ~\ref{fig4}. Notably, even with only 600 samples, AVES-DPO achieves a significant reduction in hallucination rates. We attribute this efficiency to our self-generated responses, which ensure the training data remains in-distribution, enabling the model to learn effective hallucination mitigation patterns even from small-scale datasets. Detailed quantitative results across larger dataset sizes are provided in Appendix~\ref{appendixD.1}.

\begin{figure}[t]
    \centering
    \includegraphics[width=1.0\columnwidth]{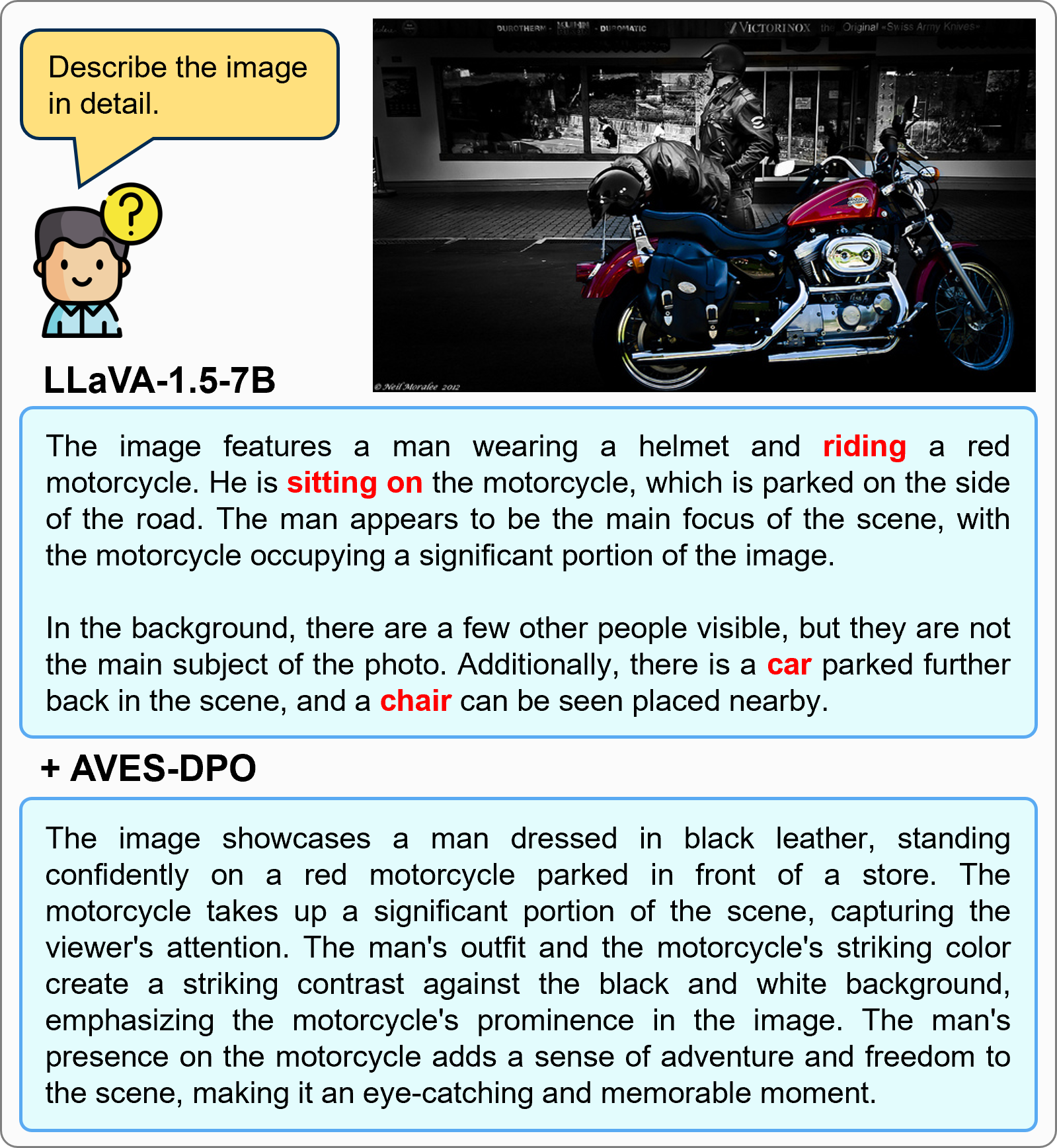}
    \caption{Qualitative comparison demonstrating the effectiveness of AVES-DPO in mitigating hallucinations.}
    \label{fig5}
\end{figure}

\noindent \textbf{Qualitative case study.}  To further validate our proposed method, we present a case study in Figure~\ref{fig5}. The baseline model exhibits severe hallucinations, generating incorrect relationships and non-existent objects such as `car' and `chair'. In contrast, AVES-DPO accurately grounds the visual content, providing a more detailed and factually correct description. This demonstrates that our approach effectively mitigates hallucination while enhancing descriptive capabilities.

%% file: latex/table/table3.tex
\begin{table*}[t]
\centering
\renewcommand{\arraystretch}{1.3}
\resizebox{1.0\textwidth}{!}{%
\begin{tabular}{l|cccccccccc|c}
\Xhline{3\arrayrulewidth}
\rowcolor[HTML]{EFEFEF} 
\multicolumn{1}{c|}{\cellcolor[HTML]{EFEFEF}\textbf{Method}} & \textbf{Existence} & \textbf{Count} & \textbf{Position} & \textbf{Color} & \textbf{Posters} & \textbf{Celebrity} & \textbf{Scene} & \textbf{Landmark} & \textbf{Artwork} & \textbf{OCR}   & \textbf{Overall} \\ \hline\hline
\textit{LLaVA-1.5-7B}                     & \textbf{195.0}     & \textbf{158.3} & 123.3             & 155.0          & 128.6            & 133.2              & 155.5          & 161.0             & \textbf{120.8}   & 125.0          & 1455.7           \\
+ POVID ~\cite{zhou2024aligning}                                   & 190.0              & \textbf{158.3} & 123.3             & 150.0          & 128.6            & 133.5              & 154.8          & 161.8             & \textbf{120.8}   & 125.0          & 1446.0           \\
+ SENTINEL ~\cite{Peng_2025_ICCV}                                & 185.0              & \textbf{158.3} & \textbf{133.3}    & \textbf{165.0} & 129.6            & 135.0              & 158.8          & \textbf{162.5}    & 120.3            & 125.0          & 1472.8           \\
\rowcolor[HTML]{EEF3FF} 
\textbf{+ AVES-DPO (Ours)}                                     & \textbf{195.0}     & 153.3          & 123.3             & 155.0          & \textbf{144.6}   & \textbf{138.8}     & \textbf{164.3} & 159.5             & 117.3            & \textbf{140.0} & \textbf{1491.1}  \\ \hline
\textit{LLaVA-1.5-13B}                    & \textbf{195.0}     & \textbf{153.3} & \textbf{120.0}    & 170.0          & \textbf{158.2}   & \textbf{168.2}     & 158.8          & 147.8             & 124.3            & \textbf{132.5} & \textbf{1528.0}  \\
+ SENTINEL ~\cite{Peng_2025_ICCV}                                & \textbf{195.0}     & 143.3          & \textbf{120.0}    & \textbf{180.0} & 156.8            & 167.1              & 161.0          & 138.0             & \textbf{126.8}   & \textbf{132.5} & 1520.4           \\
\rowcolor[HTML]{EEF3FF} 
\textbf{+ AVES-DPO (Ours)}                                     & 190.0              & 140.0          & 100.0             & 155.0          & 153.7            & 127.4              & \textbf{164.0} & \textbf{159.3}    & 101.8            & 72.5           & 1363.6           \\
\Xhline{3\arrayrulewidth}
\end{tabular}
}
\caption{Evaluation results on the MME benchmark for general multimodal perception and recognition capabilities.}
\label{tab3}
\end{table*}

%% file: latex/table/table4.tex
\begin{table}[t!]
\centering
\renewcommand{\arraystretch}{1.3}
\resizebox{1.0\columnwidth}{!}{%
\begin{tabular}{l|cc}
\Xhline{3\arrayrulewidth}

\rowcolor[HTML]{EFEFEF} 
\multicolumn{1}{c|}{\cellcolor[HTML]{EFEFEF}} & \multicolumn{2}{c}{\cellcolor[HTML]{EFEFEF}\textbf{Object-Hal}} \\ \cline{2-3} 
\rowcolor[HTML]{EFEFEF} 
\multicolumn{1}{c|}{\multirow{-2}{*}{\cellcolor[HTML]{EFEFEF}\textbf{Method}}} & \textbf{CHAIR$_S$$\downarrow$} & \textbf{CHAIR$_I$$\downarrow$} \\ \hline\hline

\textit{LLaVA-1.5-7B} & 51.4 & 14.8 \\
\rowcolor[HTML]{EEF3FF} 
\textbf{+ Hallucination Verification (Ours)} & 40.2 & 12.2 \\ \hline

\textit{LLaVA-1.5-13B}  & 50.4 & 14.3 \\
\rowcolor[HTML]{EEF3FF} 
\textbf{+ Hallucination Verification (Ours)} & 25.4 & 7.6 \\ \hline

\textit{GPT-4V} & 29.0 & 7.9 \\
\rowcolor[HTML]{EEF3FF} 
\textbf{+ Hallucination Verification (Ours)} & 23.9 & 5.6 \\
\Xhline{3\arrayrulewidth}
\end{tabular}
}
\caption{Effectiveness and robustness analysis of the proposed Hallucination Verification strategy.}
\label{tab4}
\end{table}

%% file: latex/table/table16.tex
\begin{table}[t]
\centering
\renewcommand{\arraystretch}{1.3}
\resizebox{1.0\columnwidth}{!}{%
\begin{tabular}{l|cc|ccc}
\Xhline{3\arrayrulewidth}
% === Header ===
\rowcolor[HTML]{EFEFEF} 
\multicolumn{1}{c|}{\cellcolor[HTML]{EFEFEF}} & \multicolumn{2}{c|}{\cellcolor[HTML]{EFEFEF}\textbf{Object-Hal}} & \multicolumn{3}{c}{\cellcolor[HTML]{EFEFEF}\textbf{AMBER}} \\ \cline{2-6} 
\rowcolor[HTML]{EFEFEF} 
\multicolumn{1}{c|}{\multirow{-2}{*}{\cellcolor[HTML]{EFEFEF}\textbf{Method}}} & \textbf{CHAIR$_S$$\downarrow$} & \textbf{CHAIR$_I$$\downarrow$} & \textbf{CHAIR} $\downarrow$ & \textbf{Hal Rate} $\downarrow$ & \textbf{Cog} $\downarrow$ \\ \hline\hline

% === Rows ===
\textit{LLaVA-1.5-7B} & 51.4 & 14.8 & 7.1 & 32.5 & 3.8 \\
+ GPT-4V & 38.4 & 11.4 & 7.6 & 31.7 & 3.6 \\
\rowcolor[HTML]{EEF3FF} 
\textbf{+ AVES-DPO (Ours)} & \textbf{12.2} & \textbf{3.9} & \textbf{3.3} & \textbf{12.6} & \textbf{0.9} \\
\Xhline{3\arrayrulewidth}
\end{tabular}
}
\caption{Performance comparison between external supervision from proprietary models and our self-correction approach.}
\label{tab16}
\end{table}

%% file: latex/06_conclusion.tex
In this study, we proposed AVES-DPO, a novel framework that mitigates diverse hallucinations in LVLMs via verified self-correction. By generating preference pairs derived from the models' intrinsic knowledge, our approach effectively resolves the distribution mismatch problem inherent in external feedback methods. Extensive experiments demonstrate that AVES-DPO significantly outperforms baselines on multiple benchmarks, achieving exceptional data efficiency with only 5.2k samples.

%% file: latex/09_appendix.tex
\appendix

\newpage

\section{Method Details}
\label{appendixA}

\input{latex/table/table5}

\subsection{Details for Hallucination Verification}
\label{appendixA.1}
\paragraph{Scene Graph Extraction.}
We employ the DiscoSG framework ~\cite{lin2025discosg} to parse textual responses into structured scene graphs. Specifically, we utilize flan-t5-large-VG-factual-sg as the generator and DiscoSG-Refiner-Large-t5-only as the refiner, executing two iterative refinement rounds to ensure graph quality.

\paragraph{Vocabulary Construction.} To overcome the limited category coverage of MS-COCO, we expand our vocabulary by integrating frequent concepts from the GQA dataset. The final vocabulary comprises the original 80 COCO objects augmented with 68 GQA objects, along with 38 attributes and 17 relations. Crucially, all terms are organized into semantic subtypes to facilitate the precise retrieval of correction candidates. The detailed vocabulary lists are provided in Tables ~\ref{tab5}, ~\ref{tab6}, and ~\ref{tab7}.

\paragraph{O/A/R Verification.}

\textit{(i) Object Verification.} Our pipeline relies exclusively on open-source models. For primary object detection, we utilize Grounding DINO-base and YOLOv8x-worldv2 with default configurations. To minimize interference between semantically similar queries, we implement a grouping strategy for object categories exhibiting a text embedding cosine similarity of 0.5 or higher, calculated using sentence-transformers. Furthermore, to ensure rigorous verification, scenes containing multiple instances of the same object class are excluded. Finally, discrepancies in detection results are resolved using the FP8-quantized Qwen3-VL series (30B and 32B) as a secondary verification step. The specific prompt employed to resolve these ambiguous objects is presented in Table~\ref{tab8}.

\input{latex/table/table6}
\input{latex/table/table7}

\noindent \textit{(ii) Attribute \& Relation Verification.} We employed the Qwen series models to verify the factual consistency of attributes and relationships. The specific prompt templates employed for attribute and relationship verification are presented in Table ~\ref{tab9} and Table ~\ref{tab10}, respectively. Upon detecting hallucinations, the models were instructed to generate replacements from the corresponding semantic subtypes within our predefined vocabulary. To ensure data reliability, we enforced a strict consensus mechanism. Only corrections where both models yielded identical outputs were retained, while discrepant samples were excluded to minimize noise.

\subsection{Details for Self-correction}

\paragraph{Factual Rectification.}
Guided by the error signals derived from the verification phase, the model self-refines its initial response to eliminate or rectify hallucinations. The specific prompt template utilized for this correction is presented in Table~\ref{tab11}. This approach allows for flexible sentence reconstruction while preserving the model's intrinsic linguistic style and vocabulary.

\paragraph{Detailed Enrichment.}
While factually accurate, the verified captions may become monotonous or lack sufficient visual detail. To address this and construct high-quality preferred data for DPO training, we implement a Detailed Enrichment phase. In this step, the LVLMs expand the corrected concise response by incorporating missing visual details grounded in the image. Crucially, to prevent the recurrence of errors, the model is explicitly prohibited from mentioning objects previously identified as hallucinations. The specific prompt template used for enriching the caption with visual details while avoiding hallucinations is presented in Table~\ref{tab12}. To secure a sufficient number of high-quality samples, we iterated this generation and filtering pipeline. Specifically, the process was repeated 3 times for the LLaVA-1.5-7B model and 5 times for the LLaVA-1.5-13B model.

\section{Training Details}
\label{appendixB}

\subsection{Training Dataset}

\paragraph{Dataset Statistics.} 
To ensure robust verification signals, we construct distinct training datasets for the 7B and 13B models. 
\begin{itemize}
    \item \textit{7B dataset}: comprises 5.2k samples, encompassing the verification of 19,610 objects, 2,562 attributes, and 897 relationships.
    \item \textit{13B dataset}: consists of 4.6k samples, covering 16,983 objects, 2,862 attributes, and 1,098 relationships.
\end{itemize}

\begin{figure}[t]
    \centering
    \includegraphics[width=\columnwidth]{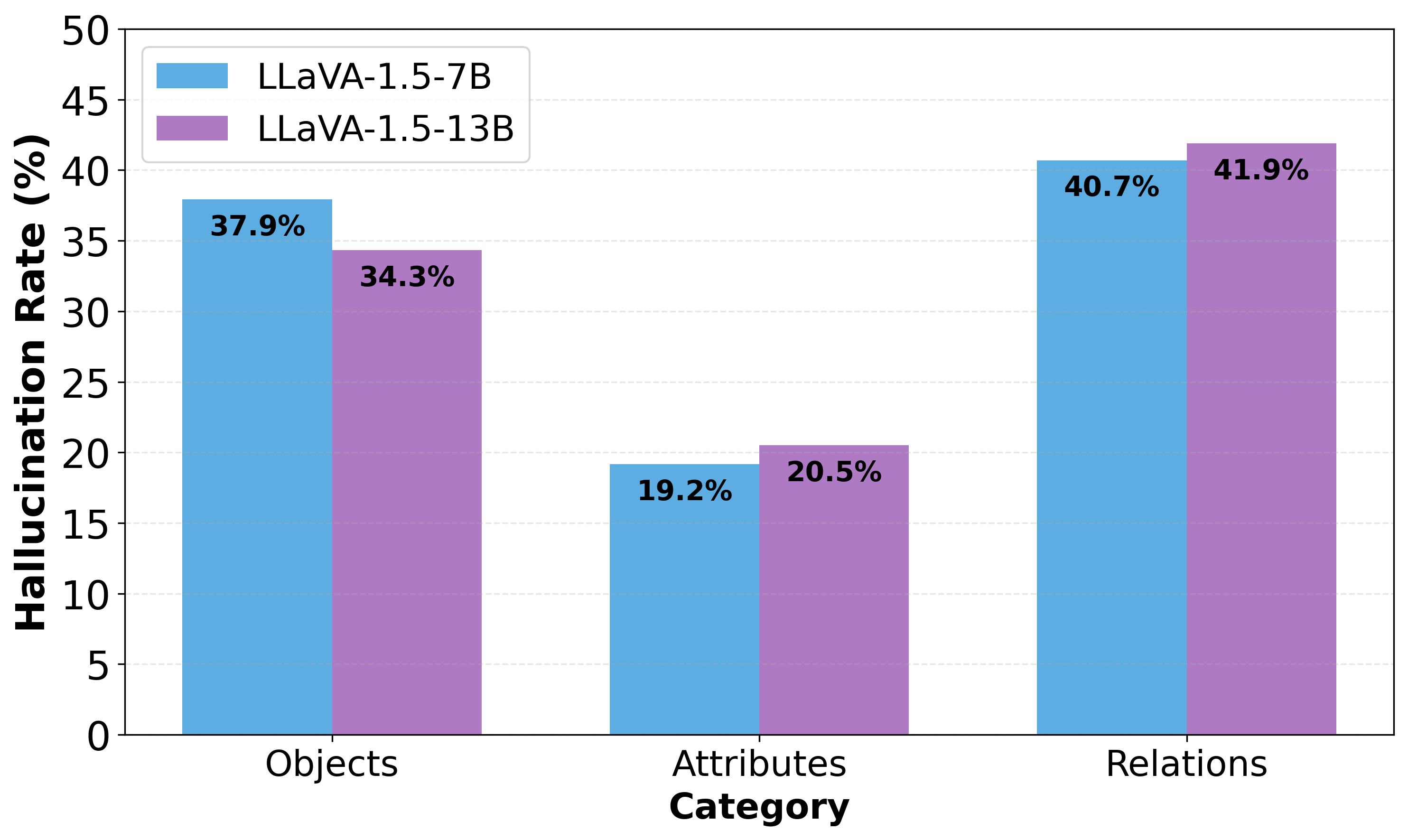}
    \caption{Comparative Analysis of Hallucination Rates. Comparison of hallucination rates across objects, attributes, and relations for LLaVA-1.5-7B and LLaVA-1.5-13B models.}
    \label{fig6}
\end{figure}

\input{latex/table/table13}

\paragraph{Hallucination Distribution.} We analyze the prevalence of hallucinations across semantic categories, as illustrated in Figure~\ref{fig6}. The distribution varies by model scale. The 7B model exhibits a higher object hallucination rate of 37.9\% compared to 34.3\% for the 13B model. Conversely, the 13B model shows increased hallucination rates for fine-grained details, recording 20.5\% for attributes and 41.9\% for relationships. Notably, relationships consistently represent the most challenging category across both model scales, highlighting the difficulty of capturing inter-object dynamics.

\subsection{Training Setup}
We fine-tuned LLaVA-1.5 (7B, 13B) using LoRA. The detailed training hyperparameters are presented in Table ~\ref{tab13}.

\input{latex/table/table14}
\input{latex/table/table15}

\section{Evaluation Details}
\label{appendixC}
\subsection{Evaluation Benchmarks}
In this section, we elaborate on the specific evaluation metrics and calculation methods for each benchmark used to measure the model's hallucination mitigation performance.

\paragraph{Object HalBench.} 
Object HalBench is widely adopted benchmark for evaluating object hallucinations in image descriptions. In this study, we generate captions using the standard image captioning prompt: ``Describe the image in detail.'' The evaluation is performed on 500 samples randomly sampled from the MS-COCO validation set, quantified by two key metrics: sentence-level (CHAIR$_S$) and object-level (CHAIR$_I$). Note that object detection within LVLMs' responses is performed using an exact matching strategy.

\begin{itemize}
    \item \textit{CHAIR$_S$ (Sentence-level Hallucination Rate):} Calculated as the ratio of captions containing at least one hallucinated object among all generated captions.
    
    % \resizebox{너비}{높이(비율유지=!)}{내용}
    \begin{center}
    \resizebox{1.0\linewidth}{!}{%
        $ \displaystyle \text{CHAIR}_S = \frac{\#\{\text{captions with hallucinated objects}\}}{\#\{\text{all captions}\}} $
    }
    \end{center}

    \item \textit{CHAIR$_I$ (Object-level Hallucination Rate):} Calculated as the ratio of hallucinated objects to the total number of objects mentioned by the model.
    
    \begin{center}
    \resizebox{0.8\linewidth}{!}{%
        $ \displaystyle \text{CHAIR}_I = \frac{\#\{\text{hallucinated objects}\}}{\#\{\text{all mentioned objects}\}} $
    }
    \end{center}
\end{itemize}

\paragraph{AMBER.} AMBER is a comprehensive benchmark encompassing both generative and discriminative tasks to analyze hallucination phenomena from multiple perspectives.

\paragraph{(Generative Task)} This task evaluates the frequency of hallucinations in model-generated responses using four metrics:
\begin{itemize}
    \item \textit{CHAIR}: Measures the accuracy by calculating the intersection ratio between the set of objects mentioned by the model and the set of objects actually present in the image.
    \item \textit{Cover}: Measures the object coverage by calculating the proportion of actual objects present in the image that are successfully mentioned by the model.
    \item \textit{Hal Rate}: Measures the proportion of responses containing hallucinations. A response is considered hallucinated if its CHAIR score exceeds 0.
    \item \textit{Cog}: Evaluates the alignment between the model's hallucinations and human cognitive tendencies. It is calculated as the probability that a pre-defined set of hallucination-prone objects appears in the generated output.
\end{itemize}

\paragraph{(Discriminative Task)} This task assesses the severity of hallucinations across six dimensions: existence, attributes, relationships, state, number, and actions. We report the model's overall performance on these aspects using Accuracy and F1 score.

\paragraph{MMHal-Bench.} 
MMHal-Bench is a Question-Answering benchmark employing GPT-4 (\texttt{gpt-4-0613}) for evaluation. Following the official protocol, model responses are scored on a scale of 0 to 6 based on factual accuracy and the hallucination presence. To calculate the final Hallucination Rate, we determine the proportion of responses that receiving a score of less than 3.

\paragraph{MME Benchmark.}
MME is a comprehensive evaluation benchmark designed to assess the perception and cognition capabilities of Multi-modal Large Language Models (MLLMs). To verify the generalization ability of our proposed method beyond simple hallucination mitigation, we conduct evaluations on subsets relevant to visual grounding and detailed understanding, including existence, count, position, and color. The total score is derived from the aggregated accuracy across these subtasks, serving as an indicator of the model's overall multimodal robustness.

\subsection{Evaluation Results}

\paragraph{Results on MMHal-Bench.} Table~\ref{tab14} summarizes the quantitative results on MMHal-Bench. For the 7B model, our method exhibits generally superior performance compared to the baselines. Notably, it demonstrates significant gains in the `Attribute', `Comparison', and `Relation' categories. This improvement aligns well with the specific types of hallucinations targeted by our proposed framework. Conversely, the LLaVA-1.5-13B model achieves the second-highest performance among the compared methods. While the improvement margin was narrower than that of the 7B model, it maintained a robust performance level, indicating competitive stability.

\paragraph{Results on POPE Benchmark.} Table ~\ref{tab15} presents the quantitative results on the POPE benchmark. Consistent with the fine-grained results observed in MMHal-Bench, the performance dynamics vary by model scale. The 7B model demonstrates clear improvements over the baseline, achieving the highest overall accuracy of 85.4 and notably excelling in the Adversarial setting. On the other hand, while the 13B model shows a marginal performance trade-off compared to the strong baseline, it maintains high stability and outperforms other alignment method such as LLaVA-RLHF.

\section{Additional Ablations and Analysis}
\label{appendixD}
\subsection{Extended Scalability Analysis}
\label{appendixD.1}
To further evaluate the scalability of our pipeline, we expanded the preference dataset up to 10.9k pairs by incorporating the GQA dataset. As shown in Table~\ref{tab17}, while our automated pipeline scales effectively, we identify approximately 5.2k pairs as a practical sweet spot. 

Beyond this scale, hallucination mitigation exhibits diminishing returns (e.g., CHAIR$_S$ decreasing only marginally from 12.2 to 10.4). More importantly, this over-optimization toward factual precision comes at the expense of descriptive richness, evidenced by a decline in the F1 score (from 72.7 to 67.0) and AMBER Cover metric (from 40.8 to 37.8). Thus, 5.2k samples establish the optimal, data-efficient balance for mitigating hallucinations while actively preserving meaningful and informative content.
\input{latex/table/table17}
\input{latex/table/table18}
\subsection{Quality Assessment of Preference Pairs}
To empirically verify that our hierarchical consensus mechanism is not bottlenecked by the performance limits of individual auxiliary models, we conducted an independent evaluation using GPT-4o on 500 randomly sampled preference pairs. As shown in Table~\ref{tab18}, our self-corrected responses achieve overwhelmingly high win rates over the initial responses across all fine-grained categories (objects, attributes, and relations). This result confirms that, despite relying on multiple open-source models, our rigorous verification pipeline successfully filters out individual model noise, consistently yielding high-quality, reliable preference pairs.

\input{latex/table/table19}
\subsection{Comprehensive Evaluation with Multimodal Judges}
To overcome the limitations of relying solely on text-based judgments for visual tasks, we expanded our assessment to include the multimodal judge-based MMHal-Bench-V~\cite{mmhal_bench_v}. As demonstrated in Table~\ref{tab19}, AVES-DPO achieves competitive performance in both hallucination reduction and overall response quality.

\input{latex/table/table20}
\subsection{Necessity of Two-Phase Object Verification}
Our two-phase object verification strategy is designed to balance precision and recall through hierarchical consensus. In the first phase, object detectors filter out explicitly hallucinated objects, forwarding only ambiguous cases to the second phase. There, two Qwen3-VL models perform strict semantic verification, effectively distinguishing actual hallucinations from simple detector failures.

To validate this mechanism, we compared the full approach against single-phase baselines on datasets of identical size. As shown in Table~\ref{tab20}, relying on a single phase falls short: Phase-1 Only yields lower descriptive richness, and Phase-2 Only suffers from significantly higher hallucination rates. By accurately resolving ambiguous cases, our two-phase strategy achieves the lowest overall hallucination rates (e.g., a CHAIR$_{S}$ of 12.2) while maintaining competitive object coverage, proving its necessity for a robust verification pipeline.

\section{AI Assistant Usage}
We have used Claude Code during the development of our research work.

\input{latex/table/table8}
\input{latex/table/table9}
\input{latex/table/table10}
\input{latex/table/table11}
\input{latex/table/table12}

%% file: latex/table/table5.tex
\begin{table*}[t]
\centering
\renewcommand{\arraystretch}{1.3}
% \resizebox 대신 tabularx를 사용하여 텍스트 너비에 맞게 자동 줄바꿈
\begin{tabularx}{\textwidth}{l|c|X}
\Xhline{3\arrayrulewidth}
\textbf{Source} & \textbf{Count} & \textbf{Object List} \\ \hline\hline
MS-COCO & 80 & person, bicycle, car, motorcycle, airplane, bus, train, truck, boat, traffic light, fire hydrant, stop sign, parking meter, bench, bird, cat, dog, horse, sheep, cow, elephant, bear, zebra, giraffe, backpack, umbrella, handbag, tie, suitcase, frisbee, skis, snowboard, sports ball, kite, baseball bat, baseball glove, skateboard, surfboard, tennis racket, bottle, wine glass, cup, fork, knife, spoon, bowl, banana, apple, sandwich, orange, broccoli, carrot, hot dog, pizza, donut, cake, chair, couch, potted plant, bed, dining table, toilet, tv, laptop, mouse, remote, keyboard, cell phone, microwave, oven, toaster, sink, refrigerator, book, clock, vase, scissors, teddy bear, hair drier, toothbrush \\ \hline
GQA & 68 & tree, building, sky, pole, window, table, door, fence, wheel, floor, jacket, hat, shoe, leaf, letter, plate, flower, bag, helmet, rock, boy, cloud, roof, cap, girl, bush, mirror, box, shelf, pillow, trunk, plant, lamp, wing, seat, house, counter, street light, glove, flag, cabinet, bike, child, container, sock, towel, mountain, basket, phone, animal, sticker, lady, license plate, cheese, wire, beach, desk, curtain, dress, tower, stone, blanket, drawer, ocean, t-shirt, television, trash can, computer \\ 
\Xhline{3\arrayrulewidth}
\end{tabularx}
\caption{List of the 148 object classes used for hallucination verification.}
\label{tab5}
\end{table*}

%% file: latex/table/table6.tex
\begin{table*}[t]
\centering
\renewcommand{\arraystretch}{1.3}
% {1.0\textwidth}로 표 너비를 고정하고, {l|X}로 두 번째 열이 남은 공간을 채우며 줄바꿈되도록 설정
\begin{tabularx}{0.85\textwidth}{l|X}
\Xhline{3\arrayrulewidth}
\textbf{Subtype} & \textbf{Attributes} \\ \hline\hline
Color & white, black, green, blue, brown, red, gray, yellow, orange, silver, pink, purple, blond, gold, beige, light brown, light blue, dark brown, cream colored, maroon, dark blue, black and white, khaki \\ \hline
Material & wooden, metal, brick \\ \hline
Posture & standing, sitting, lying, sleeping \\ \hline
Condition/State & open, closed, empty, full, wet, dry, cut, uncut \\
\Xhline{3\arrayrulewidth}
\end{tabularx}
\caption{List of the 38 attribute terms used for hallucination verification.}
\label{tab6}
\end{table*}

%% file: latex/table/table7.tex
\begin{table*}[t]
\centering
\renewcommand{\arraystretch}{1.3}
% \columnwidth로 설정하여 한쪽 단에 딱 맞게 배치 (공간 효율성)
% 두 번째 열(X)이 자동으로 줄바꿈 처리됨
\begin{tabularx}{0.85\textwidth}{l|X}
\Xhline{3\arrayrulewidth}
\textbf{Subtype} & \textbf{Relations} \\ \hline\hline
Spatial & on, under, behind, in front of, next to, on the left side of, on the right side of \\ \hline
Pose-relation & sitting on, standing on, leaning on, sitting next to, standing next to, standing in front of, standing to the right of, standing to the left of, standing in front of, standing behind \\
\Xhline{3\arrayrulewidth}
\end{tabularx}
\caption{List of the 17 relation predicates used for hallucination verification.}
\label{tab7}
\end{table*}

%% file: latex/table/table13.tex
\begin{table}[t]
\centering
\renewcommand{\arraystretch}{1.3}
\resizebox{1.0\columnwidth}{!}{%
\begin{tabular}{l|cc}
\Xhline{3\arrayrulewidth}
Setting \textbackslash\ Model & LLaVA-1.5-7B                & LLaVA-1.5-13B              \\ \hline\hline
LLM                          & Vicuna-1.5-7B               & Vicuna-1.5-13B             \\
Vision encoder               & \multicolumn{2}{c}{CLIP ViT-L 336px/14}                  \\
Projector                    & \multicolumn{2}{c}{mlp2x\_gelu}                          \\
Learning rate                & 2e-6                      & 3.2e-6                     \\
Global batch size            & 16                          & 8                          \\
Trainable parameters         & \multicolumn{2}{c}{LORA trains only LLM’s linear layers} \\
LoRA rank r                  & \multicolumn{2}{c}{128}                                  \\
LoRA alpha                   & \multicolumn{2}{c}{256}                                  \\
Optimizer                    & \multicolumn{2}{c}{AdamW}                               \\
Epochs                       & \multicolumn{2}{c}{1}                                    \\
Memory optimization          & \multicolumn{2}{c}{ZeRO stage 2}          \\\Xhline{3\arrayrulewidth}              
\end{tabular}
}
\caption{Detailed training hyperparameters for AVES-DPO.
}
\label{tab13}
\end{table}

%% file: latex/table/table14.tex
\begin{table*}[t]
\centering
\renewcommand{\arraystretch}{1.3}
\resizebox{1.0\textwidth}{!}{%
\begin{tabular}{l|c|c|cccccccc}
\Xhline{3\arrayrulewidth}
% === Header ===
\rowcolor[HTML]{EFEFEF} 
\multicolumn{1}{c|}{\cellcolor[HTML]{EFEFEF}} & \cellcolor[HTML]{EFEFEF} & \cellcolor[HTML]{EFEFEF} & \multicolumn{8}{c}{\cellcolor[HTML]{EFEFEF}\textbf{Score in Each Question Type} $\uparrow$} \\ \cline{4-11} 
\rowcolor[HTML]{EFEFEF} 
\multicolumn{1}{c|}{\multirow{-2}{*}{\cellcolor[HTML]{EFEFEF}\textbf{Method}}} & \multirow{-2}{*}{\cellcolor[HTML]{EFEFEF}\textbf{Overall} $\uparrow$} & \multirow{-2}{*}{\cellcolor[HTML]{EFEFEF}\textbf{Hall Rate} $\downarrow$} & \textbf{Attribute} & \textbf{Adv} & \textbf{Comp} & \textbf{Count} & \textbf{Rel} & \textbf{Env} & \textbf{Holistic} & \textbf{Other} \\ \hline\hline

% === 7B Models ===
\textbf{LLaVA-1.5-7B~\cite{Liu_2024_CVPR}} & {\ul 2.22} & {\ul 0.57} & 2.33 & 2.00 & 1.09 & \textbf{2.33} & 2.17 & \textbf{2.09} & \textbf{3.08} & 2.58 \\
+LLaVA-RLHF~\cite{sun-etal-2024-aligning} & 2.06 & 0.64 & \textbf{2.75} & \textbf{2.42} & 1.17 & {\ul 1.42} & \textbf{3.17} & 1.75 & 2.00 & 1.83 \\
+POVID~\cite{zhou2024aligning} & 2.18 & {\ul 0.57} & 2.58 & {\ul 2.33} & {\ul 1.25} & 1.25 & 1.92 & 1.42 & {\ul 2.92} & \textbf{3.75} \\
+SENTINEL~\cite{Peng_2025_ICCV} & 2.04 & 0.58 & 2.33 & 2.17 & 1.08 & 1.08 & 1.92 & 1.67 & 2.50 & {\ul 3.58} \\
\rowcolor[HTML]{EEF3FF} 
\textbf{+AVES-DPO (Ours)} & \textbf{2.35} & \textbf{0.53} & {\ul 2.67} & {\ul 2.33} & \textbf{2.17} & 0.83 & {\ul 2.75} & {\ul 1.83} & 2.83 & 3.42 \\ \hline

% === 13B Models ===
\textbf{LLaVA-1.5-13B~\cite{Liu_2024_CVPR}} & 2.30 & {\ul 0.54} & \textbf{3.33} & \textbf{2.83} & 1.83 & 1.58 & {\ul 1.92} & 1.42 & 2.08 & 3.42 \\
+LLaVA-RLHF~\cite{sun-etal-2024-aligning} & 2.14 & 0.66 & {\ul 2.42} & 1.67 & 1.83 & 1.08 & \textbf{2.42} & \textbf{2.33} & 1.75 & {\ul 3.58} \\
+SENTINEL~\cite{Peng_2025_ICCV} & \textbf{2.48} & \textbf{0.49} & \textbf{3.33} & 2.42 & \textbf{3.08} & \textbf{2.00} & 1.67 & 1.50 & {\ul 2.50} & 3.33 \\
\rowcolor[HTML]{EEF3FF} 
\textbf{+AVES-DPO (Ours)} & {\ul 2.36} & 0.58 & 2.33 & {\ul 2.58} & {\ul 2.42} & {\ul 1.67} & 1.58 & {\ul 1.83} & \textbf{2.83} & \textbf{3.67} \\
\Xhline{3\arrayrulewidth}
\end{tabular}
}
\caption{Fine-grained results on MMHal-Bench.}
\label{tab14}
\end{table*}

%% file: latex/table/table15.tex
\begin{table*}[t]
\centering
\renewcommand{\arraystretch}{1.3}
\resizebox{0.8\textwidth}{!}{%
\begin{tabular}{l|cc|cc|cc|cc}
\Xhline{3\arrayrulewidth}
% === Header ===
\rowcolor[HTML]{EFEFEF} 
\multicolumn{1}{c|}{\cellcolor[HTML]{EFEFEF}} & \multicolumn{2}{c|}{\cellcolor[HTML]{EFEFEF}\textbf{Random}} & \multicolumn{2}{c|}{\cellcolor[HTML]{EFEFEF}\textbf{Popular}} & \multicolumn{2}{c|}{\cellcolor[HTML]{EFEFEF}\textbf{Adversarial}} & \multicolumn{2}{c}{\cellcolor[HTML]{EFEFEF}\textbf{Overall}} \\ \cline{2-9} 
\rowcolor[HTML]{EFEFEF} 
\multicolumn{1}{c|}{\multirow{-2}{*}{\cellcolor[HTML]{EFEFEF}\textbf{Method}}} & \textbf{Acc} $\uparrow$ & \textbf{F1} $\uparrow$ & \textbf{Acc} $\uparrow$ & \textbf{F1} $\uparrow$ & \textbf{Acc} $\uparrow$ & \textbf{F1} $\uparrow$ & \textbf{Acc} $\uparrow$ & \textbf{F1} $\uparrow$ \\ \hline\hline

% === 7B Models ===
\textbf{LLaVA-1.5-7B~\cite{Liu_2024_CVPR}} & \textbf{87.3} & \textbf{86.3} & {\ul 85.6} & \textbf{84.6} & {\ul 82.9} & \textbf{82.2} & {\ul 85.3} & \textbf{84.3} \\
+LLaVA-RLHF~\cite{sun-etal-2024-aligning} & 84.2 & 82.3 & 81.9 & 80.2 & 78.9 & 77.6 & 81.7 & 80.0 \\
+SENTINEL~\cite{Peng_2025_ICCV} & {\ul 86.9} & {\ul 85.6} & 85.5 & {\ul 84.2} & 82.7 & {\ul 81.8} & 85.0 & {\ul 83.8} \\
\rowcolor[HTML]{EEF3FF} 
\textbf{+AVES-DPO (Ours)} & 86.8 & 85.0 & \textbf{85.7} & 84.0 & \textbf{83.7} & \textbf{82.2} & \textbf{85.4} & 83.7 \\ \hline

% === 13B Models ===
\textbf{LLaVA-1.5-13B~\cite{Liu_2024_CVPR}} & \textbf{90.1} & \textbf{89.6} & \textbf{88.4} & \textbf{88.1} & {\ul 84.4} & \textbf{84.6} & \textbf{87.6} & \textbf{87.4} \\
+LLaVA-RLHF~\cite{sun-etal-2024-aligning} & 84.4 & 81.9 & 82.9 & 80.5 & 81.0 & 78.8 & 82.8 & 80.4 \\
+SENTINEL~\cite{Peng_2025_ICCV} & {\ul 89.7} & {\ul 89.1} & {\ul 88.1} & {\ul 87.6} & \textbf{84.8} & \textbf{84.6} & {\ul 87.5} & {\ul 87.0} \\
\rowcolor[HTML]{EEF3FF} 
\textbf{+AVES-DPO (Ours)} & 89.2 & 88.9 & 87.3 & 87.1 & 84.0 & {\ul 84.2} & 86.8 & 86.7 \\
\Xhline{3\arrayrulewidth}
\end{tabular}
}
\caption{POPE evaluation benchmark. Accuracy denotes the accuracy of predictions. “Yes” represents the probability of the model outputting a positive answer.
}
\label{tab15}
\end{table*}

%% file: latex/table/table17.tex
\begin{table}[t]
\centering
\scriptsize
\setlength{\tabcolsep}{3.5pt}
\renewcommand{\arraystretch}{1.15}
\resizebox{\columnwidth}{!}{%
\begin{tabular}{c|ccc|cccc}
\hline
\multirow{2}{*}{\textbf{Dataset Size}} 
& \multicolumn{3}{c|}{\textbf{Object-Hal}} 
& \multicolumn{4}{c}{\textbf{AMBER}} \\
\cline{2-8}
& \textbf{CHAIR$_S\downarrow$} 
& \textbf{CHAIR$_I\downarrow$}
& \textbf{F1$\uparrow$}
& \textbf{CHAIR$\downarrow$}
& \textbf{Cover$\uparrow$}
& \textbf{Hal Rate$\downarrow$}
& \textbf{Cog$\downarrow$} \\
\hline
600   & 38.6 & 12.1 & 78.1 & 5.3 & 48.3 & 23.2 & 2.3 \\
1.3k  & 28.8 &  9.2 & 75.6 & 4.3 & 46.3 & 18.8 & 1.9 \\
2.6k  & 22.6 &  7.0 & 72.9 & 3.8 & 44.4 & 16.1 & 1.4 \\
5.2k  & 12.2 &  3.9 & 72.7 & 3.3 & 40.8 & 12.6 & 0.9 \\
8.1k  & 10.4 &  3.5 & 67.0 & 2.8 & 38.7 & 11.0 & 0.7 \\
10.9k & 10.6 &  3.5 & 65.3 & 2.8 & 37.8 & 10.1 & 0.6 \\
\hline
\end{tabular}%
}
\caption{Effect of dataset size on hallucination mitigation performance for LLaVA-1.5-7B.}
\label{tab17}
\end{table}

%% file: latex/table/table18.tex
\begin{table}[t]
\centering
\scriptsize
\setlength{\tabcolsep}{4pt}
\renewcommand{\arraystretch}{1.1}
\resizebox{\columnwidth}{!}{%
\begin{tabular}{lccc}
\hline
\textbf{Category} & \textbf{$y_w$ Win Rate $\uparrow$} & \textbf{$y_l$ Win Rate $\downarrow$} & \textbf{Tie Rate} \\
\hline
Object    & 88.3 &  9.1 & 2.6 \\
Attribute & 86.9 & 12.3 & 0.8 \\
Relation  & 85.3 & 14.3 & 0.3 \\
\hline
Overall   & 87.1 & 11.3 & 1.6 \\
\hline
\end{tabular}%
}
\caption{Win, loss, and tie rates by category.}
\label{tab18}
\end{table}

%% file: latex/table/table19.tex
\begin{table}[t]
\centering
\renewcommand{\arraystretch}{1.3}
\resizebox{0.85\columnwidth}{!}{%
\begin{tabular}{l|cc}
\Xhline{3\arrayrulewidth}
\rowcolor[HTML]{EFEFEF}
\multicolumn{1}{c|}{\cellcolor[HTML]{EFEFEF}} 
& \multicolumn{2}{c}{\cellcolor[HTML]{EFEFEF}\textbf{MMHal-Bench-V}} \\ \cline{2-3}
\rowcolor[HTML]{EFEFEF}
\multicolumn{1}{c|}{\multirow{-2}{*}{\cellcolor[HTML]{EFEFEF}\textbf{Method}}}
& \textbf{Score$\uparrow$} & \textbf{Hal Rate$\downarrow$} \\ \hline\hline

\textit{LLaVA-1.5-7B$^{\dagger}$} & 2.22 & 0.58 \\
+ LLaVA-RLHF$^{\dagger}$ ~\cite{sun-etal-2024-aligning} & 1.55 & 0.78 \\
+ POVID$^{\dagger}$ ~\cite{zhou2024aligning} & 2.30 & 0.57 \\
+ SENTINEL$^{\dagger}$ ~\cite{Peng_2025_ICCV} & \textbf{2.38} & \textbf{0.54} \\
\rowcolor[HTML]{EEF3FF}
\textbf{+ AVES-DPO (Ours)} & \underline{2.32} & \underline{0.56} \\ \hline

\textit{LLaVA-1.5-13B$^{\dagger}$} & \underline{2.47} & 0.52 \\
+ LLaVA-RLHF$^{\dagger}$ ~\cite{sun-etal-2024-aligning} & 1.35 & 0.86 \\
+ SENTINEL$^{\dagger}$ ~\cite{Peng_2025_ICCV} & \textbf{2.57} & \textbf{0.49} \\
\rowcolor[HTML]{EEF3FF}
\textbf{+ AVES-DPO (Ours)} & 2.41 & \underline{0.51} \\
\Xhline{3\arrayrulewidth}
\end{tabular}
}
\caption{Comparison on MMHal-Bench-V. For baseline algorithms with available official checkpoints, we re-evaluate the models, and these results are marked with $^{\dagger}$.}
\label{tab19}
\end{table}

%% file: latex/table/table20.tex
\begin{table}[t]
\centering
\scriptsize
\setlength{\tabcolsep}{4pt}
\renewcommand{\arraystretch}{1.15}
\resizebox{\columnwidth}{!}{%
\begin{tabular}{l|cc|cccc}
\hline
\multirow{2}{*}{\textbf{Verification}} 
& \multicolumn{2}{c|}{\textbf{Object-Hal}} 
& \multicolumn{4}{c}{\textbf{AMBER}} \\
\cline{2-7}
& \textbf{CHAIR$_S\downarrow$} 
& \textbf{CHAIR$_I\downarrow$} 
& \textbf{CHAIR}$\downarrow$ 
& \textbf{Cover}$\uparrow$ 
& \textbf{Hal Rate}$\downarrow$ 
& \textbf{Cog}$\downarrow$ \\
\hline
Two-Phase    & 12.2 & 3.9 & 3.3 & 40.8 & 12.6 & 0.9 \\
Phase-1 Only & 16.6 & 4.8 & 3.5 & 40.9 & 12.8 & 1.0 \\
Phase-2 Only & 19.8 & 6.1 & 3.9 & 44.7 & 17.0 & 1.2 \\
\hline
\end{tabular}%
}
\caption{Ablation results of verification strategies.}
\label{tab20}
\end{table}

%% file: latex/table/table8.tex
\begin{table*}[t]
\centering
\renewcommand{\arraystretch}{1.2}
% textwidth로 설정하여 논문 폭에 꽉 차게 배치 (1단/2단 모두 호환)
\begin{tabularx}{1.0\textwidth}{X}
\Xhline{3\arrayrulewidth}
\rowcolor[HTML]{EFEFEF} 
\textbf{System Prompt for Object Verification} \\ \hline\hline
You are a precise vision assistant analyzing objects in the image. \\
\\
\textbf{\#\#\# CRITICAL INSTRUCTIONS} \\
1. Focus on the target object specified in the task. \\
2. Determine if the target object exists and is correctly identified in the image. \\
\\
\textbf{\#\#\# DECISION GUIDELINES} \\
- \textbf{CORRECT}: The target object is clearly present and correctly identified in the image. \\
- \textbf{INCORRECT}: The target object is NOT present in the image. \\
- \textbf{UNCLEAR}: The object is too blurry, too small, too dark to identify, or it is ambiguous whether it matches the target object. \\
\\
\textbf{\#\#\# YOUR TASK} \\
Based on the provided image, verify the target object below. \\
\\
Object: ``\{object\_name\}'' \\
\\
Output: \\ \Xhline{3\arrayrulewidth}
\end{tabularx}
\caption{The prompt template used for verifying the existence of objects in the image.}
\label{tab8}
\end{table*}

%% file: latex/table/table9.tex
\begin{table*}[t]
\centering
\renewcommand{\arraystretch}{1.2}
\begin{tabularx}{1.0\textwidth}{X}
\Xhline{3\arrayrulewidth}
\rowcolor[HTML]{EFEFEF} 
\textbf{System Prompt for Attribute Verification} \\ \hline\hline
You are a precise vision assistant analyzing object attributes in the image. \\
\\
\textbf{\#\#\# CRITICAL INSTRUCTIONS} \\
1. Focus on the target object specified in the task. \\
2. Verify if the target attributes accurately describe the target object. \\
3. If an attribute is clearly INCORRECT: \\
\quad - TRY to select a correction from the provided ``ATTRIBUTE TYPE GUIDANCE'' if a suitable option exists. \\
\quad - If NO suitable option exists in the guidance, mark as INCORRECT without providing a correction. \\
\\
\textbf{\#\#\# DECISION GUIDELINES} \\
- \textbf{CORRECT}: The attribute visually matches the target object. \\
- \textbf{INCORRECT}: The attribute does NOT match. Provide a correction from the ``ATTRIBUTE TYPE GUIDANCE'' list if possible. If no suitable correction exists, leave the correction empty. \\
- \textbf{UNCLEAR}: The attribute cannot be determined due to occlusion, ambiguity, or lack of visual evidence. Do NOT provide a correction. \\
\\
\textbf{\#\#\# YOUR TASK} \\
Based on the provided image, verify the target attributes below. \\
\\
Object: ``\{object\_name\}'' \\
Attributes to verify: \{attributes\} \\
ATTRIBUTE TYPE GUIDANCE: \\
\{attribute\_guidance\} \\
\\
Output: \\ \Xhline{3\arrayrulewidth}
\end{tabularx}
\caption{The prompt template used for verifying attributes of the target object.}
\label{tab9}
\end{table*}

%% file: latex/table/table10.tex
\begin{table*}[t]
\centering
\renewcommand{\arraystretch}{1.2}
\begin{tabularx}{1.0\textwidth}{X}
\Xhline{3\arrayrulewidth}
\rowcolor[HTML]{EFEFEF} 
\textbf{System Prompt for Relation Verification} \\ \hline\hline
You are a precise vision assistant analyzing the relationship between two objects in the image. \\
\\
\textbf{\#\#\# CRITICAL INSTRUCTIONS} \\
1. Focus STRICTLY on the relationship between the subject and object specified in the task. \\
2. Verify if the target relation semantically and accurately describes the visual relationship. \\
3. If the relation is clearly INCORRECT: \\
\quad - TRY to select a correction from the provided ``RELATION TYPE GUIDANCE'' if a suitable option exists. \\
\quad - If NO suitable option exists in the guidance, mark as INCORRECT without providing a correction. \\
\\
\textbf{\#\#\# DECISION GUIDELINES} \\
- \textbf{CORRECT}: The relation accurately and semantically describes the visual relationship between the objects. Consider semantic equivalence. If the relation meaning matches what you see, mark it as CORRECT. \\
- \textbf{INCORRECT}: The relation is clearly wrong and contradicts what you see in the image. Provide a correction from the ``RELATION TYPE GUIDANCE'' list if possible. If no suitable correction exists, leave the correction empty. \\
- \textbf{UNCLEAR}: The relationship cannot be determined due to occlusion, ambiguity, or lack of visual evidence. Do NOT provide a correction. \\
\\
\textbf{\#\#\# YOUR TASK} \\
Based on the provided image, verify the target relation below. \\
\\
Subject: ``\{subject\_name\}'' \\
Relation: ``\{relation\}'' \\
Object: ``\{object\_name\}'' \\
Relation to verify: ``\{subject\_name\} \{relation\} \{object\_name\}'' \\
RELATION TYPE GUIDANCE: \\
\{relation\_guidance\} \\
\\
Output: \\ \Xhline{3\arrayrulewidth}
\end{tabularx}
\caption{The prompt template used for verifying relationships between objects.}
\label{tab10}
\end{table*}

%% file: latex/table/table11.tex
\begin{table*}[t]
\centering
\renewcommand{\arraystretch}{1.2}
\begin{tabularx}{1.0\textwidth}{X}
\Xhline{3\arrayrulewidth}
\rowcolor[HTML]{EFEFEF} 
\textbf{System Prompt for Caption Correction} \\ \hline\hline
You are an intelligent text editor. Fix the errors in the caption based on the ISSUES. \\
\\
\textbf{\#\#\# RULES} \\
1. Remove objects: Remove ONLY the object mention and its related phrase completely. Do NOT delete an entire sentence unless the whole sentence is only about removed objects. If an object listed in ISSUES is not found in the caption, IGNORE it. \\
2. Replace: When 'A' -> 'B' is provided, REPLACE 'A' with 'B'. \\
3. Remove: Delete ONLY the specified adjective or relation phrase entirely. \\
4. Grammar \& Style: Fix ONLY the items listed in ISSUES. NEVER add sentences stating what is missing. Output ONLY the final fixed caption directly. \\
\\
\textbf{\#\#\# YOUR TASK} \\
ORIGINAL: \{original\_caption\} \\
ISSUES: \\
\{hallucination\_info\} \\
FIXED: \\ \Xhline{3\arrayrulewidth}
\end{tabularx}
\caption{The prompt template used for correcting captions based on identified issues.}
\label{tab11}
\end{table*}

%% file: latex/table/table12.tex
\begin{table*}[t]
\centering
\renewcommand{\arraystretch}{1.2}
\begin{tabularx}{1.0\textwidth}{X}
\hline\Xhline{3\arrayrulewidth}\rowcolor[HTML]{EFEFEF} 
\textbf{System Prompt for Caption Enrichment} \\ \hline\hline
You are a precise vision assistant. Your task is to enrich the provided `Basic Description' into a detailed, natural paragraph based on the image. \\
\\
\textbf{\#\#\# RULES} \\
1. You must keep all the existing facts from the Basic Description exactly as they are, maintaining the original sentence structure as much as possible. \\
2. You should actively identify and include other objects or details that are clearly visible in the image but are missing from the Basic Description. \\
3. You must strictly utilize only visual evidence. Do not infer emotions or intentions. \\
4. You must combine the original facts and new visual details into a single, cohesive, and natural-sounding paragraph. \\
5. Describe ONLY what is visible. NEVER mention what is missing (e.g., strictly avoid phrases like ``there is no'', ``not present'', ``does not contain'', ``not visible'', ``no visible''). \\
\\
\textbf{\#\#\# Basic Description} \\
\{refined\_caption\} \\
\\
\textbf{\#\#\# CRITICAL WARNING (Negative Constraints)} \\
The following objects have been confirmed as NOT present in the image. You must NEVER mention or describe them: \\
- \{hallucinated\_objects\} \\
\\
\textbf{\#\#\# Enriched Description} \\ \Xhline{3\arrayrulewidth}
\end{tabularx}
\caption{The prompt template used for enriching the caption with visual details while avoiding hallucinations.}
\label{tab12}
\end{table*}